\documentclass[runningheads]{llncs}
\usepackage{microtype}
\usepackage{graphicx}
\usepackage{comment}
\usepackage{amsmath,amssymb} 
\usepackage{pifont}
\usepackage{color}
\usepackage{algorithmic}
\usepackage{booktabs}
\usepackage{multirow}
\usepackage[ruled, vlined, linesnumbered]{algorithm2e}
\usepackage[normalem]{ulem}
\usepackage{bm}
\usepackage{caption}
\usepackage{subcaption}
\usepackage{hyperref}
\usepackage{wrapfig}
\usepackage[title,toc,titletoc,header]{appendix}

\useunder{\uline}{\ul}{}
\def\argmin{\mathop{\rm argmin}\limits}

\def\Minimize{\mathop{\rm minimize}\limits}

\def\st{\mathop{\rm subject\ to}}

\def\ourmethod{MSuNAS}
\def\ourmodel{NSGANetV2}
\begin{document}
\pagestyle{headings}
\mainmatter
\title{\ourmodel{}: Evolutionary Multi-Objective Surrogate-Assisted Neural Architecture Search} 

\titlerunning{Evolutionary Multi-Obj. Surrogate-Assisted NAS}
\author{Zhichao Lu \and Kalyanmoy Deb \and Erik Goodman \and Wolfgang Banzhaf \and\\Vishnu Naresh Boddeti}
\authorrunning{Z. Lu et al.}
\institute{Michigan State University, East Lansing, MI 48824, USA
\email{\{luzhicha,kdeb,goodman,banzhafw,vishnu\}@msu.edu}}
\maketitle


\begin{abstract}
In this paper, we propose an efficient NAS algorithm for generating task-specific models that are competitive under multiple competing objectives. It comprises of two surrogates, one at the architecture level to improve sample efficiency and one at the weights level, through a supernet, to improve gradient descent training efficiency. On standard benchmark datasets (C10, C100, ImageNet), the resulting models, dubbed NSGANetV2, either match or outperform models from existing approaches with the search being orders of magnitude more sample efficient. Furthermore, we demonstrate the effectiveness and versatility of the proposed method on six diverse non-standard datasets, e.g. STL-10, Flowers102, Oxford Pets, FGVC Aircrafts etc. In all cases, NSGANetV2s improve the state-of-the-art (under mobile setting), suggesting that NAS can be a viable alternative to conventional transfer learning approaches in handling diverse scenarios such as small-scale or fine-grained datasets. Code is available at \url{https://github.com/mikelzc1990/nsganetv2}.

\end{abstract}

\section{Introduction}
Neural networks have achieved remarkable performance on large scale supervised learning tasks in computer vision. A majority of this progress was achieved by architectures designed manually by skilled practitioners. Neural Architecture Search (NAS) \cite{nasnet} attempts to automate this process to find good architectures for a given dataset. This promise has led to tremendous improvements in convolutional neural network architectures, in terms of predictive performance, computational complexity and model size on standard large-scale image classification benchmarks such as ImageNet~\cite{imagenet}, CIFAR-10~\cite{cifar}, CIFAR-100~\cite{cifar} etc. However, the utility of these developments, has so far eluded more widespread and practical applications. These are cases where one wishes to use NAS to obtain high-performance models on custom non-standard datasets, optimizing possibly multiple competing objectives, and to do so without the steep computation burden of existing NAS methods.

The goal of NAS is to obtain both the optimal architecture and its associated optimal weights. The key barrier to realizing the full potential of NAS is the nature of its formulation. NAS is typically treated as a bi-level optimization problem, where an inner optimization loops over the weights of the network for a given architecture, while the outer optimization loops over the network architecture itself. The computational challenge of solving this problem stems from both the upper and lower level optimization. Learning the optimal weights of the network in the lower level necessitates costly iterations of stochastic gradient descent. Similarly, exhaustively searching the optimal architecture is prohibitive due to the discrete nature of the architecture description, size of search space and our desire to optimize multiple, possibly competing, objectives. Mitigating both of these challenges explicitly and simultaneously is the goal of this paper.

Many approaches have been proposed to improve the efficiency of NAS algorithms, both in terms of the upper level and the lower level. A majority of them focuses on the lower level, including weight sharing~\cite{baker2017accelerating,enas,darts}, proxy models \cite{nasnet,amoebanet}, coarse training \cite{mnasnet}, etc. But these approaches still have to sample, explicitly or implicitly, a large number of architectures to evaluate in the upper level. In contrast, there is relatively little focus on improving the sample efficiency of the upper level optimization. A few recent approaches~\cite{PNAS,chamnet} adopt surrogates that predict the lower level performance with the goal of navigating the upper level search space efficiently. However, these surrogate predictive models are still very sample inefficient since they are learned in an offline stage by first sampling a large number of architectures that require full lower level optimization.

\begin{figure}[t]
    \centering
    \begin{subfigure}{0.95\textwidth}
    \centering
    \includegraphics[width=\textwidth{}]{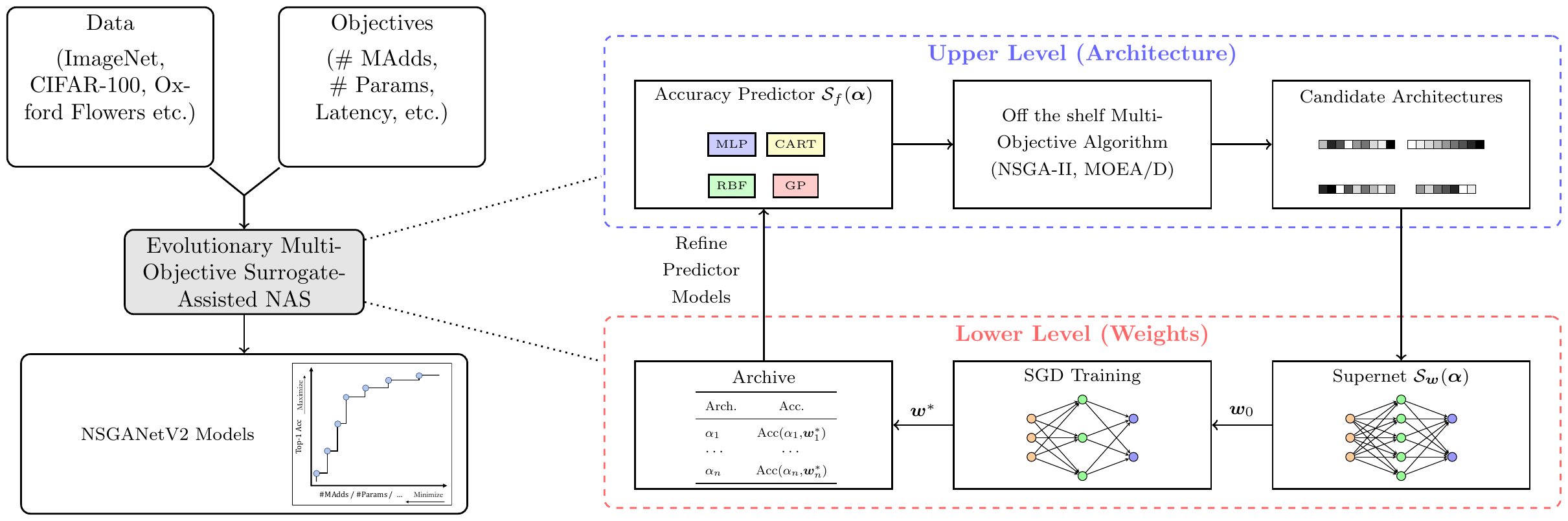}
    \end{subfigure}
\caption{\textbf{Overview:} Given a dataset and objectives, \ourmethod{} obtains a task-specific set of models that are competitive in all objectives with high search efficiency. It comprises of two surrogates, one at the upper level to improve sample efficiency and one at the lower level, through a supernet, to improve weight learning efficiency.\label{fig:overview}}
\end{figure}

In this paper, we propose a practically efficient NAS algorithm, by adopting explicit surrogate models simultaneously at both the upper and the lower level. Our lower level surrogate adopts a fine-tuning approach, where the initial weights for fine-tuning are obtained by a supernet model, such as \cite{baker2017accelerating,smash,onceforall}. Our upper level surrogate adopts an online learning algorithm, that focuses on architectures in the search space that are close to the current trade-off front, as opposed to a random/uniform set of architectures used in the offline surrogate approaches~\cite{chamnet,dppnet,PNAS}. Our online surrogate significantly improves the sample efficiency of the upper level optimization problem in comparison to the offline surrogates. For instance, OnceForAll~\cite{onceforall} and PNAS~\cite{PNAS} sample 16,000 and 1,160\footnote{Estimate from \# of models evaluated by PNAS, actual sample size is not reported.} architectures, respectively, to learn the upper level surrogate. In contrast, we only have to sample 350 architectures to obtain a model with similar performance.

An overview of our approach is shown in Fig.\ref{fig:overview}. We refer to the proposed NAS algorithm as \ourmethod{} and the resulting architectures as \ourmodel{}. Our method is designed to provide a set of high-performance models on a custom dataset (large or small scale, multi-class or fine-grained) while optimizing possibly multiple objectives of interest. Our key contributions are:

\vspace{3pt}
\noindent\textbf{-} An alternative approach to solve the bi-level NAS problem, i.e., simultaneously optimizing the architecture and learn the optimal model weights. However, instead of gradient based relaxations (e.g., DARTS), we advocate for surrogate models. Overall, given a dataset and a set of objectives to optimize, \ourmethod{} can design custom neural network architectures as efficiently as DARTS but with higher performance and extends to multiple, possibly competing objectives.

\vspace{3pt}
\noindent\textbf{-} A simple, yet highly effective, online surrogate model for the upper level optimization in NAS, resulting in a significant increase in sampling efficiency over other surrogate-based approaches.

\vspace{3pt}
\noindent\textbf{-} Scalability and practicality of \ourmethod{} on many datasets corresponding to different scenarios. These include standard datasets like ImageNet, CIFAR-10 and CIFAR-100, and six non-standard datasets like CINIC-10~\cite{cinic10} (multi-class), STL-10~\cite{stl-10}(small scale mutli-class), Oxford Flowers102~\cite{flowers102}(small scale fine-grained) etc. Under mobile settings ($\leq$ 600M MAdds), \ourmethod{} leads to SOTA performance.

\section{Related Work}
\begin{table}[!t]
\caption{Comparison of Existing NAS methods}
\label{tab:related-work}
\centering
\resizebox{0.85\textwidth}{!}{%
\begin{tabular}{@{\hspace{2mm}}l|c|ccc|c@{\hspace{2mm}}}
\toprule
Methods \hspace{1mm} & \hspace{1mm} \begin{tabular}[c]{@{}c@{}}Search Method \\ \end{tabular} \hspace{1mm} & \hspace{1mm} \begin{tabular}[c]{@{}c@{}} Performance \\ Prediction \end{tabular} & \hspace{1mm} \begin{tabular}[c]{@{}c@{}} Weight \\ Sharing \end{tabular} & \hspace{1mm} \begin{tabular}[c]{@{}c@{}} Multiple\\ Objective \end{tabular} \hspace{1mm} & \begin{tabular}[c]{@{}c@{}}Dataset Searched \\ \end{tabular} \\ \midrule
NASNet \cite{nasnet} & RL &  &  &  & C10 \\
ENAS \cite{enas} & RL &  & \checkmark &  & C10 \\
PNAS \cite{PNAS} & SBMO & \checkmark &  &  & C10 \\
DPP-Net \cite{dppnet} & SBMO & \checkmark &  & \checkmark & C10 \\
DARTS \cite{darts} & Gradient &  & \checkmark &  & C10 \\
LEMONADE \cite{LEMONADE} & EA &  & \checkmark & \checkmark & C10, C100 \\
ProxylessNAS \cite{proxylessnas} & RL + gradient &  & \checkmark & \checkmark & C10, ImageNet \\
MnasNet \cite{mnasnet} & RL &  &  & \checkmark & ImageNet \\
ChamNet \cite{chamnet} & EA & \checkmark &  & \checkmark & ImageNet \\
MobileNetV3 \cite{mobilenetv3} & RL + expert &  &  & \checkmark & ImageNet \\ \midrule
\textbf{\ourmethod{} (ours)} & EA & \checkmark & \checkmark & \checkmark & \begin{tabular}[c]{@{}c@{}}C10, C100, ImageNet,\\Pets, STL-10, Aircraft,\\DTD, CINIC-10, Flowers102\end{tabular} \\ \bottomrule
\end{tabular}%
}
\end{table}

\noindent\textbf{Lower Level Surrogate:} Existing approaches~\cite{enas,smash,darts,luo2018neural} primarily focus on mitigating the computational overhead induced by SGD-based weight optimization in the lower level, as this process needs to be repeated for every architecture sampled by a NAS method in the upper level. A common theme among these methods involves training a supernet which contains all searchable architectures as its sub-networks. During search, accuracy using the weights inherited from the supernet becomes the metric to select architectures. However, completely relying on supernet as a substitute of actual weight optimization for evaluating candidate architectures is unreliable. Numerous studies \cite{li2019random,xie2019exploring,Yu2020Evaluating} reported a weak correlation between the performance of the searched architectures (predicted by weight sharing) and the ones trained from scratch (using SGD) during the evaluation phase. \ourmethod{} instead uses the weights inherited from the supernet only as an initialization to the lower level optimization. Such a fine-tuning process affords the computation benefit of the supernet, while at the same time improving the correlation in the performance of the weights initialized from the supernet and those trained from scratch.

\vspace{5pt}
\noindent\textbf{Upper Level Surrogate:} MetaQNN \cite{baker2017accelerating} uses surrogate models to predict the final accuracy of candidate architectures (as a time-series prediction) from the first 25\% of the learning curve from SGD training. PNAS \cite{PNAS} uses a surrogate model to predict the top-1 accuracy of architectures with an additional branch added to the cell structure that are repeatedly stacked together. Fundamentally, both of these approaches seek to extrapolate rather than interpolate the performance of the architecture using the surrogates. Consequently, as we show later in the paper, the rank-order between the predicted accuracy and the true accuracy is very low\footnote{In the Appendix \ref{sec:surrogate} we show that better rank-order correlation at the search stage ultimately leads to finding better performing architectures.} (0.476). OnceForAll \cite{onceforall} also uses a surrogate model to predict accuracy from architecture encoding. However, the surrogate model is trained offline for the entire search space, thereby needing a large number of samples for learning (16K samples -$>$ 2 GPU-days -$>$ 2x search cost of DARTS for just constructing the surrogate model). Instead of using uniformly sampled architectures and their validation accuracy to train the surrogate model to approximate the entire landscape, ChamNet \cite{chamnet} trains many architectures through full lower level optimization and selects only 300 samples with high accuracy with diverse efficiency (FLOPs, Latency, Energy) to train a surrogate model offline. In contrast, \ourmethod{} learns a surrogate model in an online fashion only on the samples that are close to the current trade-off front as we explore the search space. The online learning approach significantly improves the sample efficiency of our search, since we only need lower level optimization (full or surrogate assisted) for the samples near the current Pareto front.

\vspace{5pt}
\noindent\textbf{Multi-Objective NAS:} Approaches that consider more than one objective to optimize the architecture can be categorized into two groups: (i) scalarization, and (ii) population based approaches. The former include, ProxylessNAS \cite{proxylessnas}, MnasNet \cite{mnasnet}, FBNet \cite{fbnet}, and MobileNetV3 \cite{mobilenetv3} which use a scalarized objective that encourages high accuracy and penalizes compute inefficiency at the same time, e.g., maximize $Acc * (Latency / Target)^{-0.07}$. These methods require a pre-defined preference weighting of the importance of different objectives before the search, which typically requires a numbers of trials. Methods in the latter category include \cite{NSGANet,LEMONADE,dppnet,chu2019fairnas,muxconv} and aim to approximate the entire Pareto-efficient frontier simultaneously. These approaches rely on heuristics (e.g., EA) to efficiently navigate the search space, which allows practitioners to visualize the trade-off between the objectives and to choose a suitable network \emph{a posteriori} to the search. \ourmethod{} falls in the latter category using surrogate models to mitigate the computational overhead.

\section{Proposed Approach \label{sec:approach}}

The neural architecture search problem for a target dataset $\mathcal{D} = \{\mathcal{D}_{trn}, \mathcal{D}_{vld}, \mathcal{D}_{tst}\}$ can be formulated as the following bilevel optimization problem \cite{bilevel},
\begin{equation}
\begin{aligned}
\Minimize & \hspace{3mm} \mathbf{F}(\bm{\alpha}) = \big(f_1(\bm{\alpha}; \bm{w}^*(\bm{\alpha})), \ldots, f_k(\bm{\alpha}; \bm{w}^*(\bm{\alpha})), f_{k+1}(\bm{\alpha}), \ldots, f_m(\bm{\alpha})\big)^T, \\
\st  & \hspace{3mm} \bm{w}^*(\bm{\alpha}) \in \argmin~\mathcal{L}(\bm{w};\bm{\alpha}), \\
     & \hspace{3mm} \bm{\alpha} \in \mathbf{\Omega}_{\alpha}, \hspace{3mm} \bm{w} \in \mathbf{\Omega}_{w},
\end{aligned}
\label{def:bilevel}
\end{equation}
where the upper level variable $\bm{\alpha}$ defines a candidate CNN architecture, and the lower level variable $\bm{w}(\bm{\alpha})$ defines the associated weights. $\mathcal{L}(\bm{w};\bm{\alpha})$ denotes the cross-entropy loss on the training data $\mathcal{D}_{trn}$ for a given architecture $\bm{\alpha}$. $\mathbf{F}: \mathbf{\Omega} \rightarrow \mathbb{R}^m$ constitutes $m$ desired objectives. These objectives can be further divided into two groups, where the first group ($f_1$ to $f_k$) consists of objectives that depend on both the architecture and the weights---e.g., predictive performance on validation data $\mathcal{D}_{vld}$, robustness to adversarial attack, etc. The other group ($f_{k+1}$ to $f_m$) consists of objectives that only depend on the architecture---e.g., number of parameters, floating point operations, latency etc.

\subsection{Search Space}
\begin{figure}[t]
    \centering
    \includegraphics[width=0.85\textwidth{}]{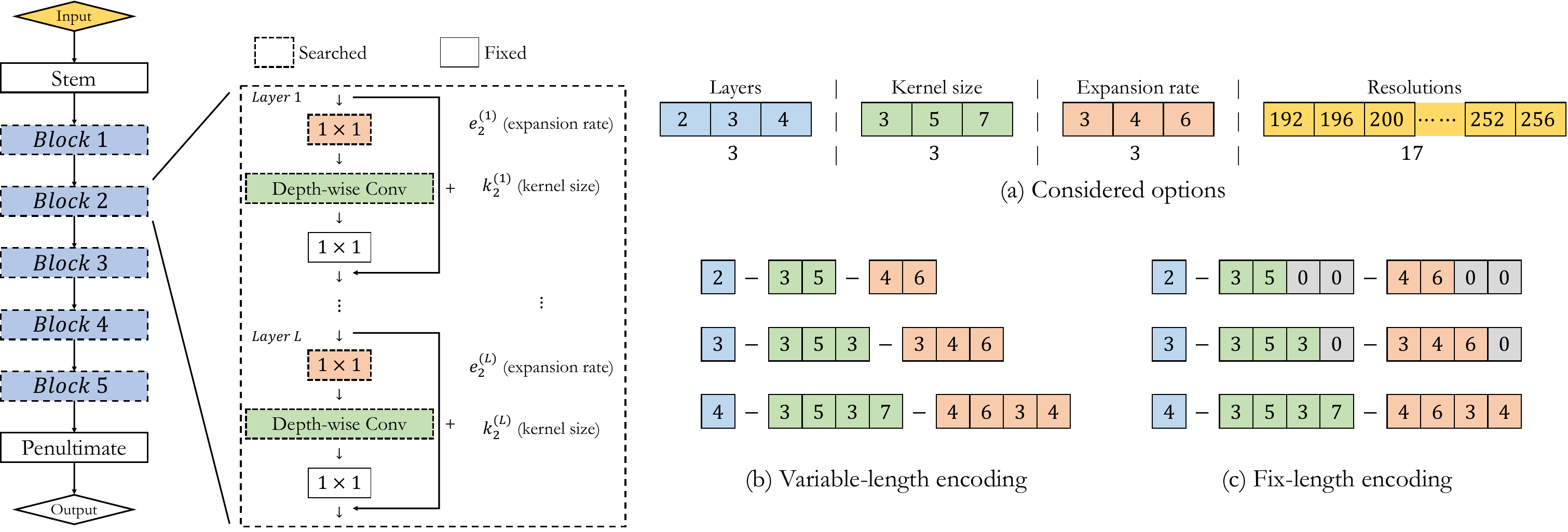}
    \caption{\textbf{Search Space}: A candidate architecture comprises five computational blocks. Parameters we search for include image resolution, number of layers ($L$) in each block and the expansion rate ($e$) and the kernel size ($k$) in each layer.
    \label{fig:search_space}}
\end{figure}

\ourmethod{} searches over four important dimensions of convolutional neural networks (CNNs), including depth (\# of layers), width (\# of channels), kernel size and input resolution. Following previous works \cite{mnasnet,mobilenetv3,onceforall}, we decompose a CNN architecture into five sequentially connected blocks, with gradually reduced feature map size and increased number of channels. In each block, we search over the number of layers, where only the first layer uses stride 2 if the feature map size decreases, and we allow each block to have minimum of two and maximum of four layers. Every layer adopts the inverted bottleneck structure \cite{mobilenetv2} and we search over the expansion rate in the first $1\times1$ convolution and the kernel size of the depth-wise separable convolution. Additionally, we allow the input image size to range from 192 to 256. We use an integer string to encode these architectural choices, and we pad zeros to the strings of architectures that have fewer layers so that we have a fixed-length encoding. A pictorial overview of this search space and encoding is shown in Fig.~\ref{fig:search_space}.

\subsection{Overall Algorithm Description}
\begin{figure}[!t]
    \centering
    \begin{minipage}{.43\textwidth}
        \begin{center}
        \scalebox{0.85}{
            \small{
            \begin{algorithm}[H]
            \SetAlgoLined
            \SetKwInOut{Input}{Input}
            \SetKwInOut{Output}{Output}
            \SetKwFor{For}{for}{do}{end for}
            \Input{$\mathcal{SS}$ (search space),\\ $\mathcal{S}_w$ (supernet),\\ $\mathcal{C}$ (complexity obj),\\ $N$ (initial samples),\\ $K$ (max. iterations).}
                $\mathcal{A}$ $\leftarrow$ $\emptyset$\;
                \While {$i < N$}{
                    $\alpha$ $\leftarrow$ sample($\mathcal{SS}$) \\
                    $w_o \leftarrow \mathcal{S}_w(\alpha)$ \\
                    $acc \leftarrow \mbox{SGD}(\alpha, w_o)$ \\
                    $\mathcal{A}$ $\leftarrow$ $\mathcal{A}$ $\cup$ $(\alpha, acc)$ \\
                }
                \While {$j < K$}{
                    $\mathcal{S}_f$ $\leftarrow$ construct from $\mathcal{A}$ // \scriptsize{(MLP / CART / RBF / GP)}\small\\
                    $\bm{\tilde{\alpha}}$ $\leftarrow$ NSGA-II($\mathcal{S}_f$, $\mathcal{C}$)\\
                    $\bm{\alpha}$ $\leftarrow$ subset from $\bm{\tilde{\alpha}}$\\
                    \For {$\alpha$ in $\bm{\alpha}$}{
                        $w_o \leftarrow \mathcal{S}_w(\alpha)$ \\
                        $acc \leftarrow \mbox{SGD}(\alpha, w_o)$ \\
                        $\mathcal{A}$ $\leftarrow$ $\mathcal{A}$ $\cup$ $(\alpha, acc)$ \\
                    }
                }
            \textbf{Return} NDsort($\mathcal{A}$).
            \caption{\ourmethod{}\label{algo:nas}}
            \end{algorithm}
        }}
        \end{center}
    \end{minipage}\hfill
    \begin{minipage}{0.57\textwidth}
        \centering
        \includegraphics[width=0.98\textwidth{}]{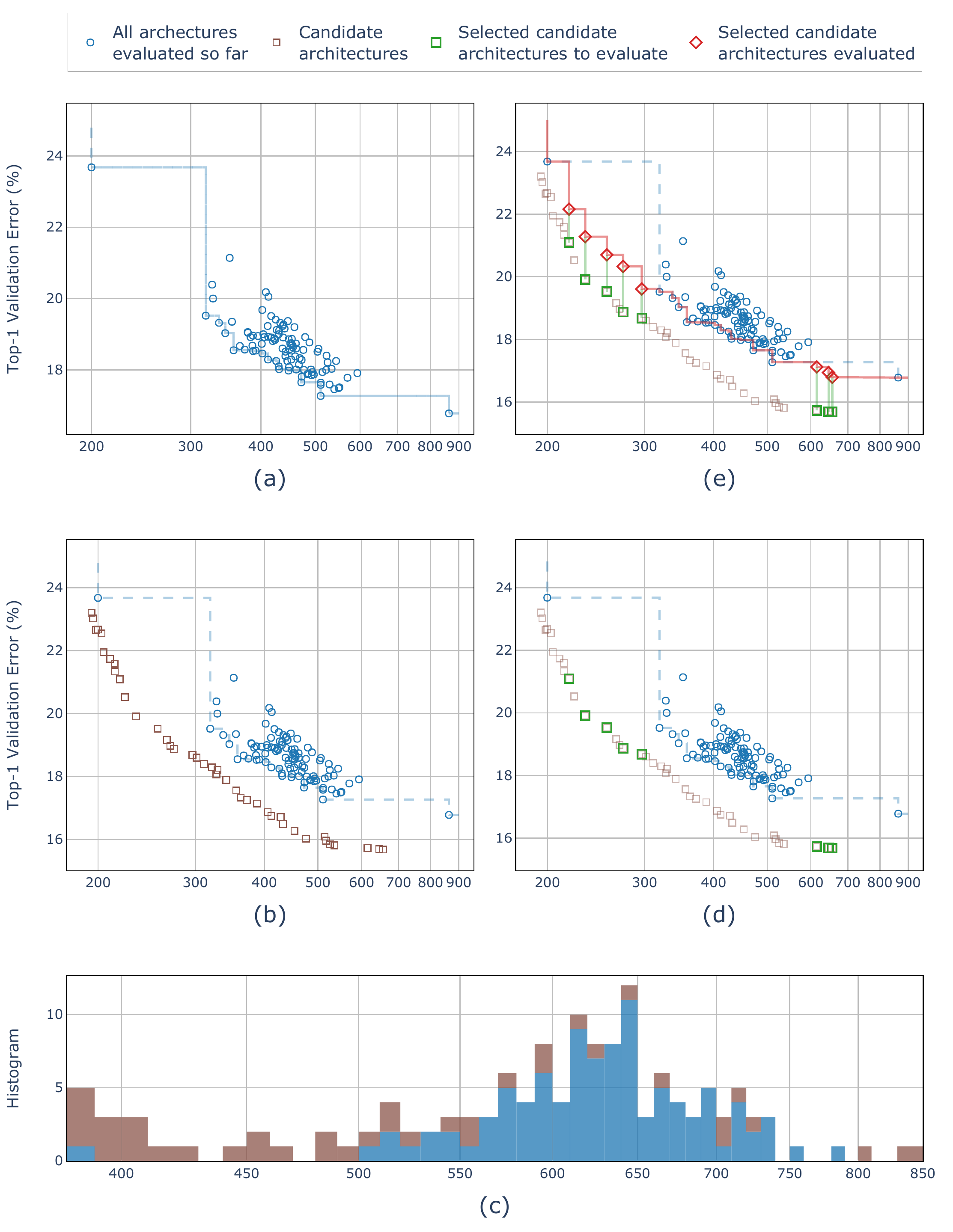}
    \end{minipage}
    \caption{A sample run of \ourmethod{} on ImageNet: In each iteration, accuracy-prediction surrogate models $\mathcal{S}_f$ are constructed from an archive of previously evaluated architectures (a). New candidate architectures \big(brown boxes in (b)\big) are obtained by solving the auxiliary single-level multi-objective problem $\tilde{F} = \{\mathcal{S}_f, \mathcal{C}\}$ (line 10 in Algo~\ref{algo:nas}). A subset of the candidate architectures is chosen to diversify the Pareto front (c) - (d). The selected candidate architectures are then evaluated and added to the archive (e). At the conclusion of search, we report the non-dominated architectures from the archive. The x-axis in all sub-figures is \#MAdds.
    \label{fig:algo}}
\end{figure}
The problem in Eq.~\ref{def:bilevel} poses two main computational bottlenecks for conventional bi-level optimization methods. First, the lower level problem of learning the optimal weights $\bm{w}^*(\bm{\alpha})$ for a given architecture $\bm{\alpha}$ involves a prolonged training process---e.g., one complete SGD training on ImageNet dataset takes two days on an 8-GPU server. Second, even though there exist techniques like weight-sharing to bypass the gradient-descent-based weight learning process, extensively sampling architectures at the upper level can still render the overall process computationally prohibitive, e.g., 10,000 evaluations on ImageNet take 24 GPU hours, and for methods like NASNet, AmoebaNet that require more than 20,000 samples, it still requires days to complete the search even with weight-sharing.

Algorithm~\ref{algo:nas} and Fig.~\ref{fig:algo} show the pseudocode and corresponding steps from a sample run of \ourmethod{} on ImageNet, respectively. To overcome the aforementioned bottlenecks, we use surrogate models at both upper and lower levels to make our NAS algorithm practically useful for a variety of datasets and objectives. At the upper level, we construct a surrogate model that predicts the top-1 accuracy from integer strings that encode architectures. Previous approaches \cite{chamnet,ae-cnn+e2epp,onceforall} that also used surrogate-modeling of the accuracy follow an offline approach, where the accuracy predictor is built from samples collected separately prior to the architecture search and not refined during the search. We argue that such a process makes the search outcome highly dependent on the initial training samples. As an alternative, we propose to model and refine the accuracy predictor iteratively in an online manner during the search. In particular, we start with an accuracy predictor constructed from only a limited number of architectures sampled randomly from the search space. We then use a standard multi-objective algorithm (NSGA-II~\cite{nsga2}, in our case) to search using the constructed accuracy predictor along with other objectives that are also of interest to the user. We then evaluate the outcome architectures from NSGA-II and refine the accuracy predictor model with these architectures as new training samples. We repeat this process for a pre-specified number of iterations and output the non-dominated solutions from the pool of evaluated architectures.

\subsection{Speeding Up Upper Level Optimization}
Recall that the nested nature of the bi-level problem makes the upper level optimization computationally very expensive, as every upper level function evaluation requires another optimization at the lower level. Hence, to improve the efficiency of our approach at the upper level, we focus on reducing the number of architectures that we send to the lower level for learning optimal weights. To achieve this goal, we need a surrogate model to predict the accuracy of an architecture before we actually train it. There are two desired properties of such a predictor: (1) high rank-order correlation between predicted and true performance; and (2) sample efficient such that the required number of architectures to be trained through SGD are minimized for constructing the predictor.

We first collected four different surrogate models for accuracy prediction from the literature, namely, Multi Layer Perceptron (MLP) \cite{PNAS}, Classification And Regression Trees (CART) \cite{ae-cnn+e2epp}, Radial Basis Function (RBF) \cite{baker2017accelerating} and Gaussian Process (GP) \cite{chamnet}. From our ablation study, we observed that no one surrogate model is consistently better than others in terms of the above two criteria on all datasets (see section~\ref{subsec:surrogate}). Hence, we propose a selection mechanism, dubbed Adaptive Switching (AS), which constructs all four types of surrogate models at every iteration and adaptively selects the best model via cross-validation.

With the accuracy predictor selected by AS, we apply the NSGA-II algorithm to simultaneously optimize for both accuracy (predicted) and other objectives of interest to the user (line 10 in Algorithm~\ref{algo:nas}). For the purpose of illustration, we assume that the user is interested in optimizing \#MAdds as the second objective. At the conclusion of the NSGA-II search, a set of non-dominated architectures is output, see Fig.~\ref{fig:algo}(b). Often times, we cannot afford to train all architectures in the set. To select a subset, we first select the architecture with highest predicted accuracy. Then we project all other architecture candidates to the \#MAdds axis, and pick the remaining architectures from the sparse regions that help in extending the Pareto frontier to diverse \#MAdds regimes, see Fig.~\ref{fig:algo}(c) - (d). The architectures from the chosen subset are then sent to the lower level for SGD training. We finally add these architectures to the training samples to refine our accuracy predictor models and proceed to next iteration, see Fig.~\ref{fig:algo}(e).

\subsection{Speeding Up Lower Level Optimization}
To further improve the search efficiency of the proposed algorithm, we adopt the widely-used weight-sharing technique \cite{smash,luo2018neural,atomnas}. First, we need a supernet such that all searchable architectures are sub-networks of it. We construct such a supernet by taking the searched architectural hyperparameters at their maximum values, i.e., with four layers in each of the five blocks, with expansion ratio set to 6 and kernel size set to 7 in each layer (See Fig.~\ref{fig:search_space}). Then we follow the progressive shrinking algorithm \cite{onceforall} to train the supernet. This process is executed once before the architecture search. The weights inherited from the trained supernet are used as a warm-start for the gradient descent algorithm during architecture search.

\section{Experiments and Results}
In this section, we evaluate the surrogate predictor, the search efficiency and the obtained architectures on CIFAR-10 \cite{cifar}, CIFAR-100 \cite{cifar}, and ImageNet \cite{imagenet}.

\subsection{Performance of the Surrogate Predictors\label{subsec:surrogate}}
\begin{figure}[t]
    \begin{subfigure}[b]{0.98\textwidth}
    \centering
    \includegraphics[width=\textwidth{}]{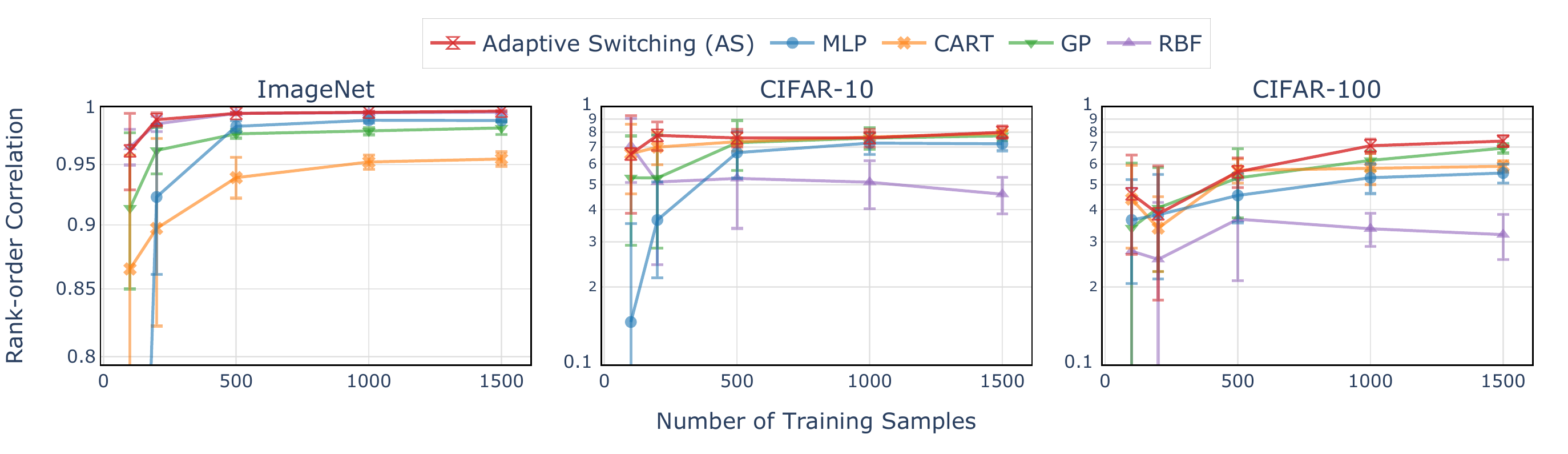}
    \end{subfigure}\\
    \begin{subfigure}[b]{0.32\textwidth}
    \includegraphics[width=\textwidth{}]{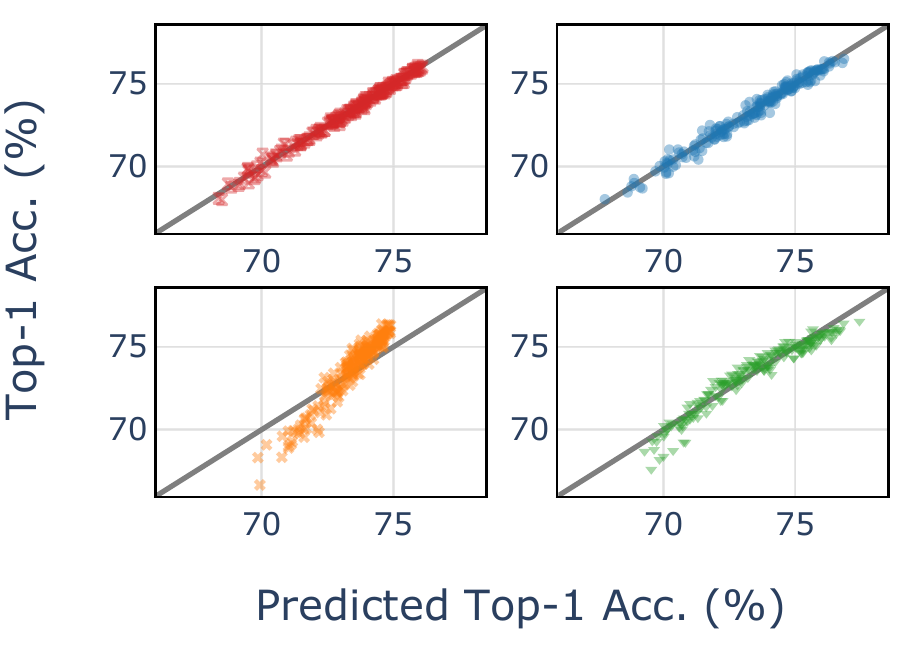}
    \end{subfigure}
    \begin{subfigure}[b]{0.32\textwidth}
    \includegraphics[width=\textwidth{}]{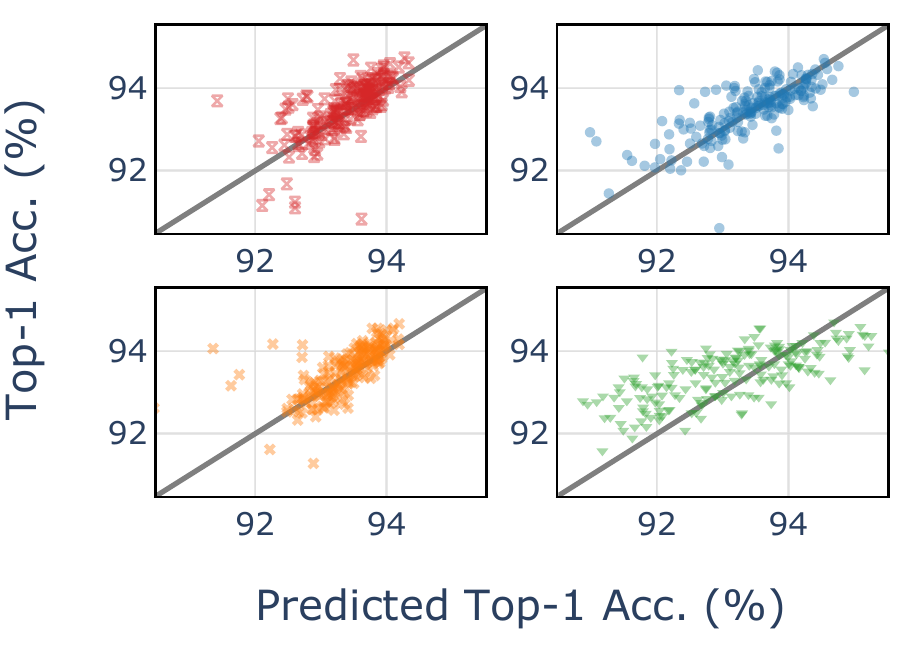}
    \end{subfigure}
    \begin{subfigure}[b]{0.32\textwidth}
    \includegraphics[width=\textwidth{}]{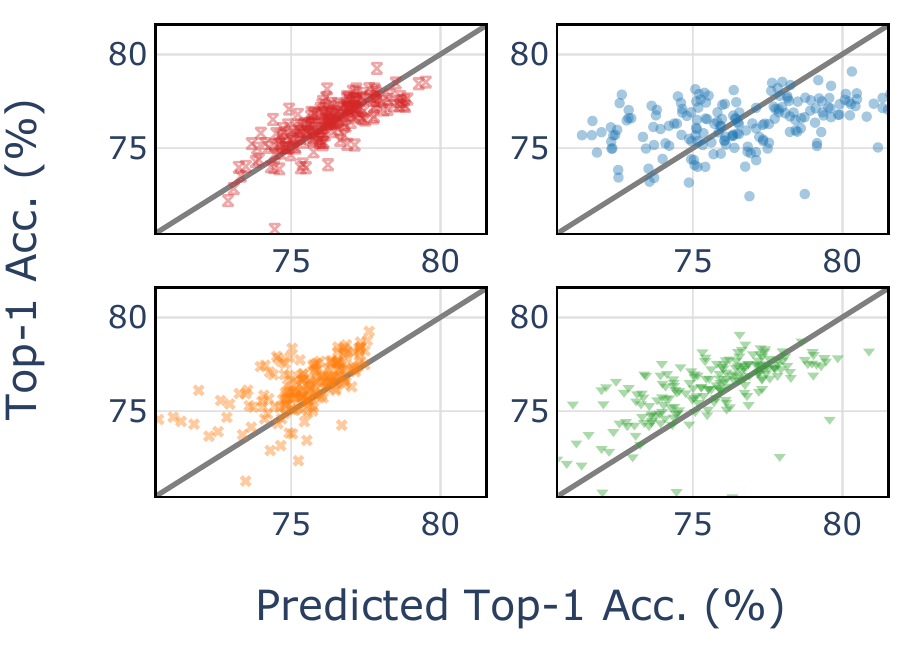}
    \end{subfigure}
    \caption{Comparing the relative prediction performance of the proposed Adaptive Switching (AS) method to the existing four surrogate models. Top row compares Spearman rank-order correlation coefficient as number of training samples increases. Bottom row visualizes the true vs. predicted accuracy under 500 training samples (RBF method is omitted to conserve space).
    \label{fig:surrogate}}
\end{figure}

To evaluate the effectiveness of the considered surrogate models, we uniformly sample 2,000 architectures from our search space, and train them using SGD for 150 epochs on each of the three datasets and record their accuracy on 5,000 held-out images from the training set. We then fit surrogate models with different number of samples randomly selected from the 2,000 collected. We repeat the process for 10 trials to compare the mean and standard deviation of the rank-order correlation between the predicted and true accuracy, see Fig.~\ref{fig:surrogate}. In general, we observe that no single surrogate model consistently outperforms the others on all three datasets. Hence, at every iteration, we adopt an Adaptive Switching (AS) routine that compares the four surrogate models and chooses the best based on 10-fold cross-validation. It is evident from Fig.~\ref{fig:surrogate} that AS works better than any one of the four surrogate models alone on all three datasets. The construction time of the AS is negligible (relatively to the search cost).

\subsection{Search Efficiency}
\begin{table}[hbt]
\caption{Comparing the relative search efficiency of \ourmethod{} to other single-objective methods: ``\#Model'' is the total number of architectures evaluated during search, ``\#Epochs'' is the number of epochs used to train each architecture during search. $^\dagger$ and $^\ddagger$ denote training epochs with and without a supernet to warm-start the weights, respectively.
\label{tab:search_cost_breakdown}}
\centering
\resizebox{0.9\textwidth}{!}{%
\begin{tabular}{c|lc|cc|cc|cc@{\hspace{2mm}}}
\toprule
\multicolumn{1}{l|}{} & \hspace{1mm} Method & \hspace{1mm} Type \hspace{1mm} & \hspace{1mm} Top1 Acc. & \hspace{1mm} \#MAdds \hspace{1mm} & \hspace{1mm} \#Model & \hspace{1mm} Speedup \hspace{1mm} & \hspace{1mm} \#Epochs & \hspace{1mm} Speedup \\ \midrule
\multirow{4}{*}{CIFAR-10} & NASNet-A \cite{nasnet} & RL & 97.4\% & 569M & 20,000 & 57x & 20 & up to 4x \\
 & AmoebaNet-B \cite{amoebanet} & EA & 97.5\% & 555M & 27,000 & 77x & 25 & up to 5x \\
 & PNASNet-5 \cite{PNAS} & SMBO & 96.6\% & 588M & 1,160 & 3.3x & 20 & up to 4x \\
 & \textbf{\ourmethod{}(ours)} & EA & 98.4\% & 468M & 350 & 1x & 5$^\dagger$ / 20$^\ddagger$ & 1x \\ \midrule
\multicolumn{1}{l|}{\multirow{3}{*}{ImageNet}} & MnasNet-A \cite{mnasnet} & RL & 75.2\% & 312M & 8,000 & 23x & 5 & up to 5x \\
\multicolumn{1}{l|}{} & OnceForAll \cite{onceforall} & EA & 76.0\% & 230M & 16,000 & 46x & 0 & - \\
\multicolumn{1}{l|}{} & \textbf{\ourmethod{}(ours)} & EA & 75.9\% & 225M & 350 & 1x & 0$^\dagger$ / 5$^\ddagger$ & 1x \\ \bottomrule
\end{tabular}%
}
\end{table}

\begin{figure}[!thb]
    \begin{subfigure}[b]{0.5\textwidth}
    \centering
    \includegraphics[width=.64\textwidth{}]{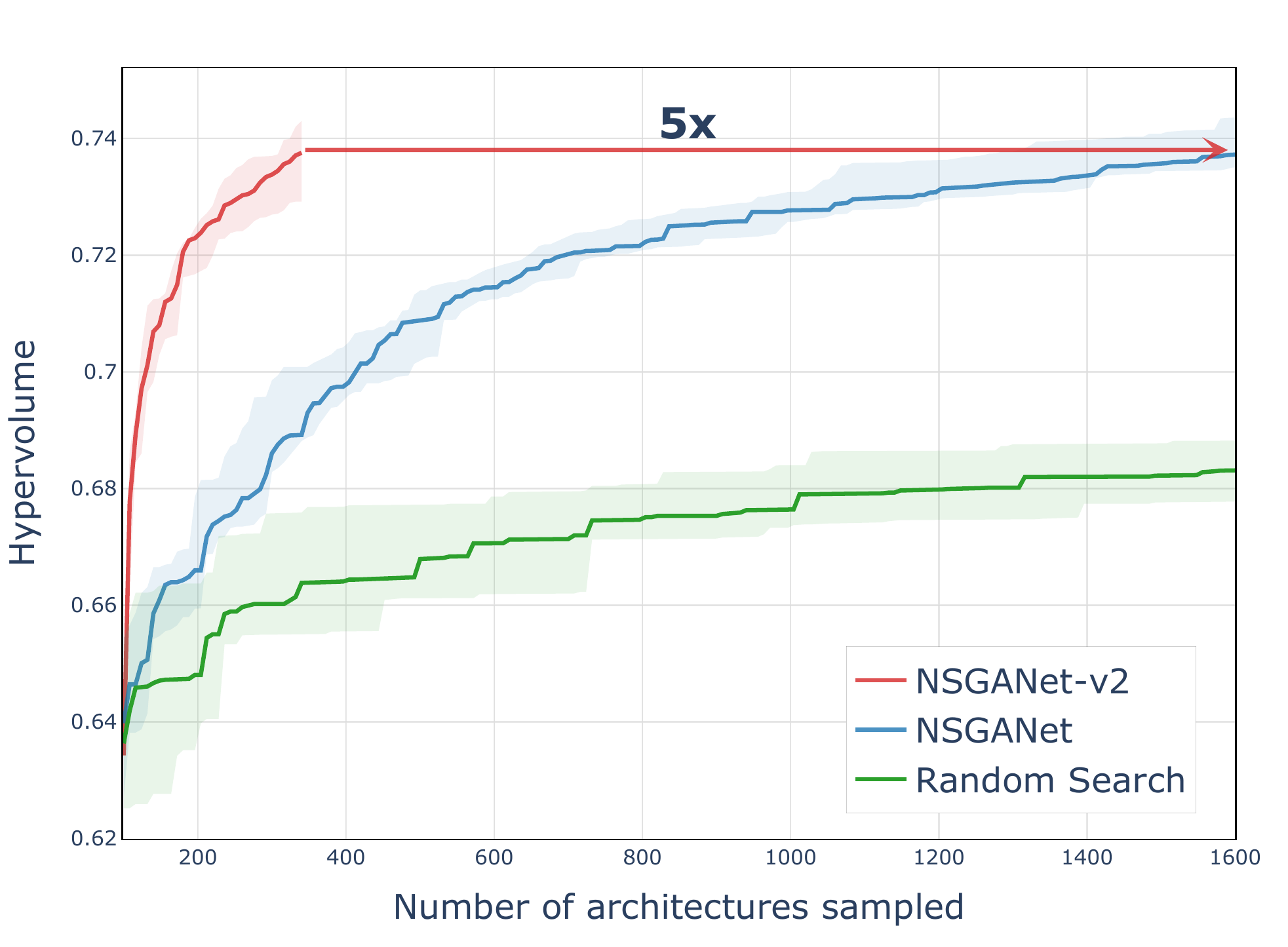}
    \includegraphics[width=.34\textwidth{}]{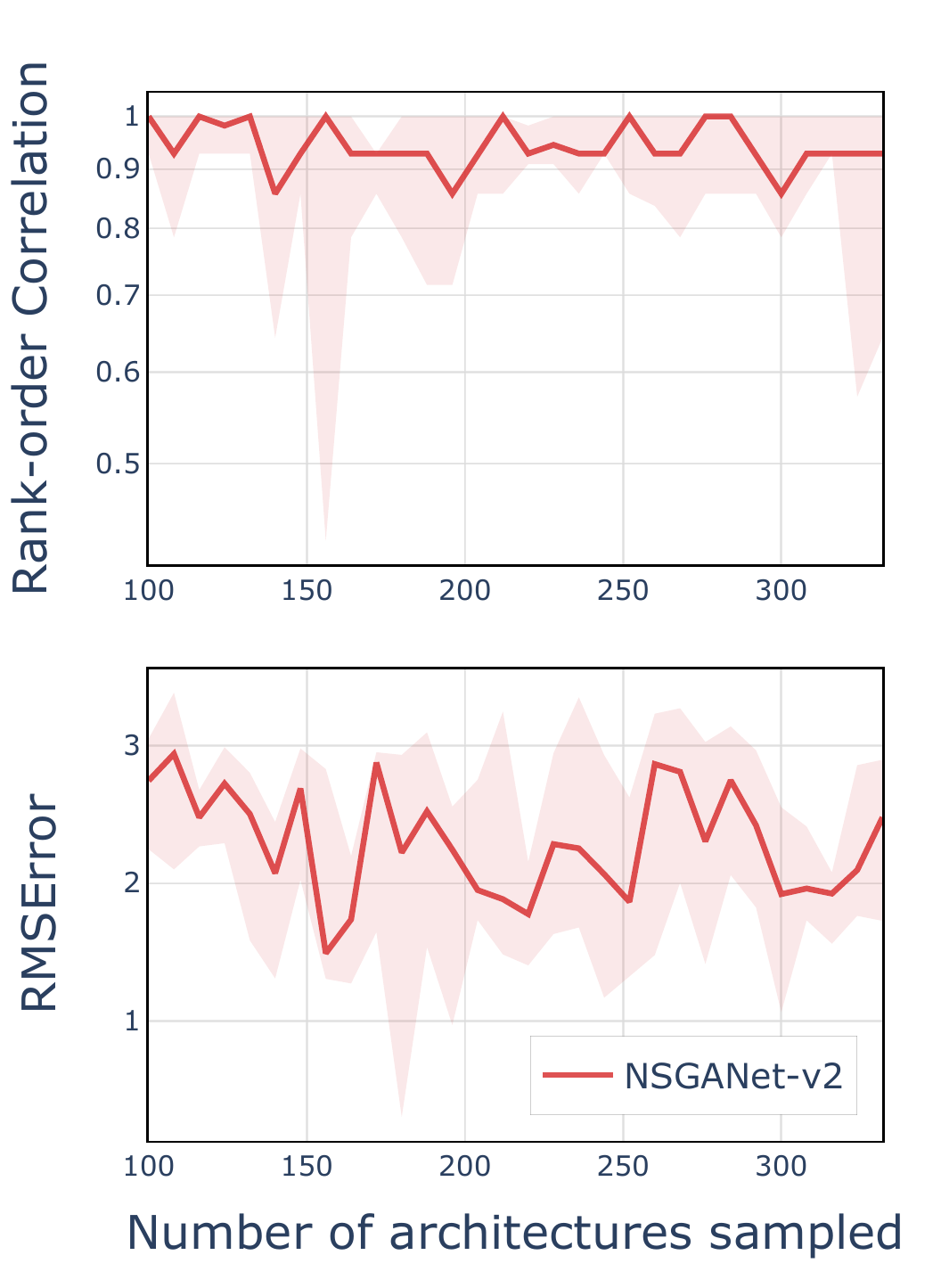}
    \caption{ImageNet
    \label{fig:imagenet_search}}
    \end{subfigure}\hfill
    \begin{subfigure}[b]{0.5\textwidth}
    \centering
    \includegraphics[width=.64\textwidth{}]{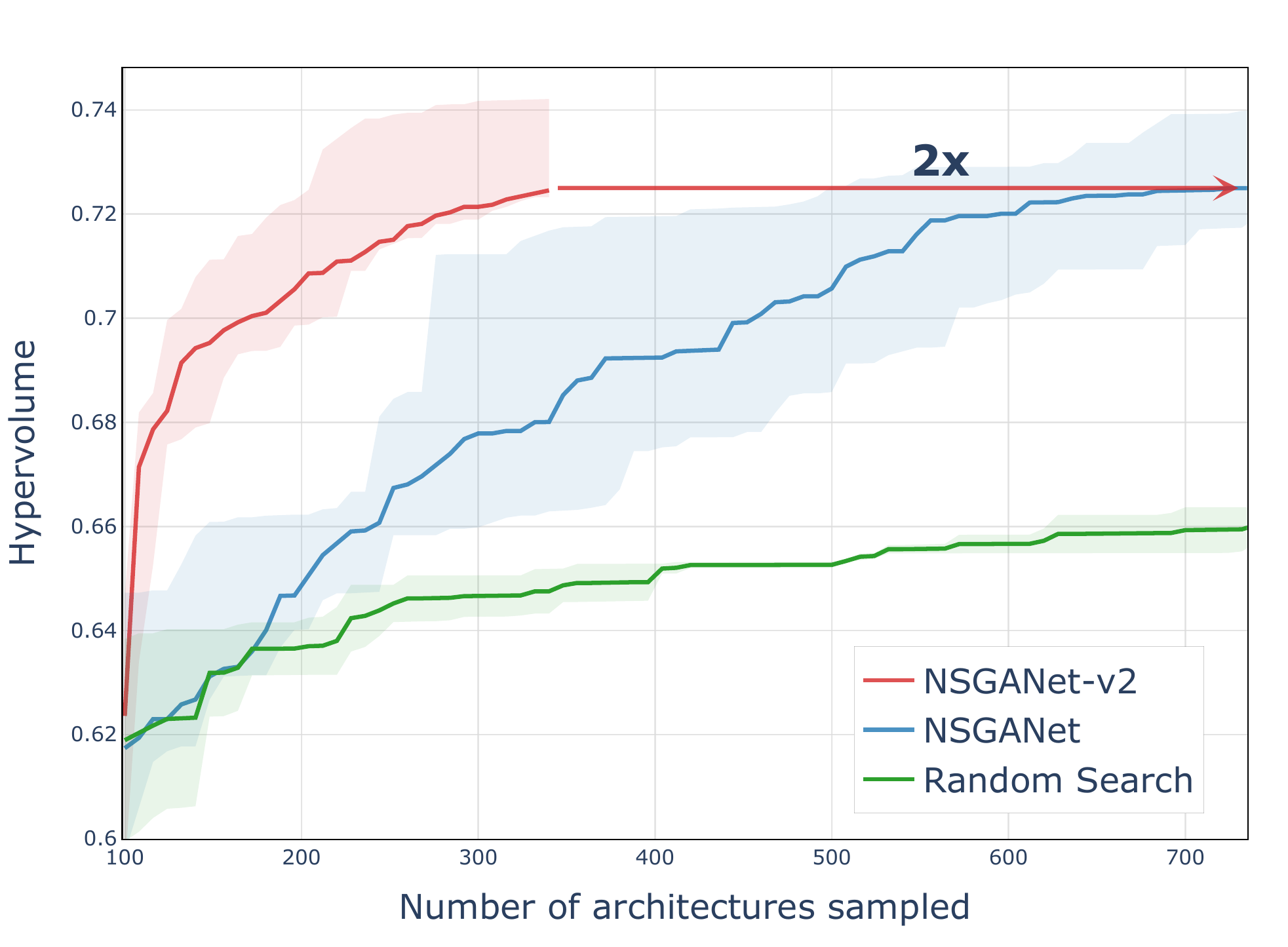}
    \includegraphics[width=.34\textwidth{}]{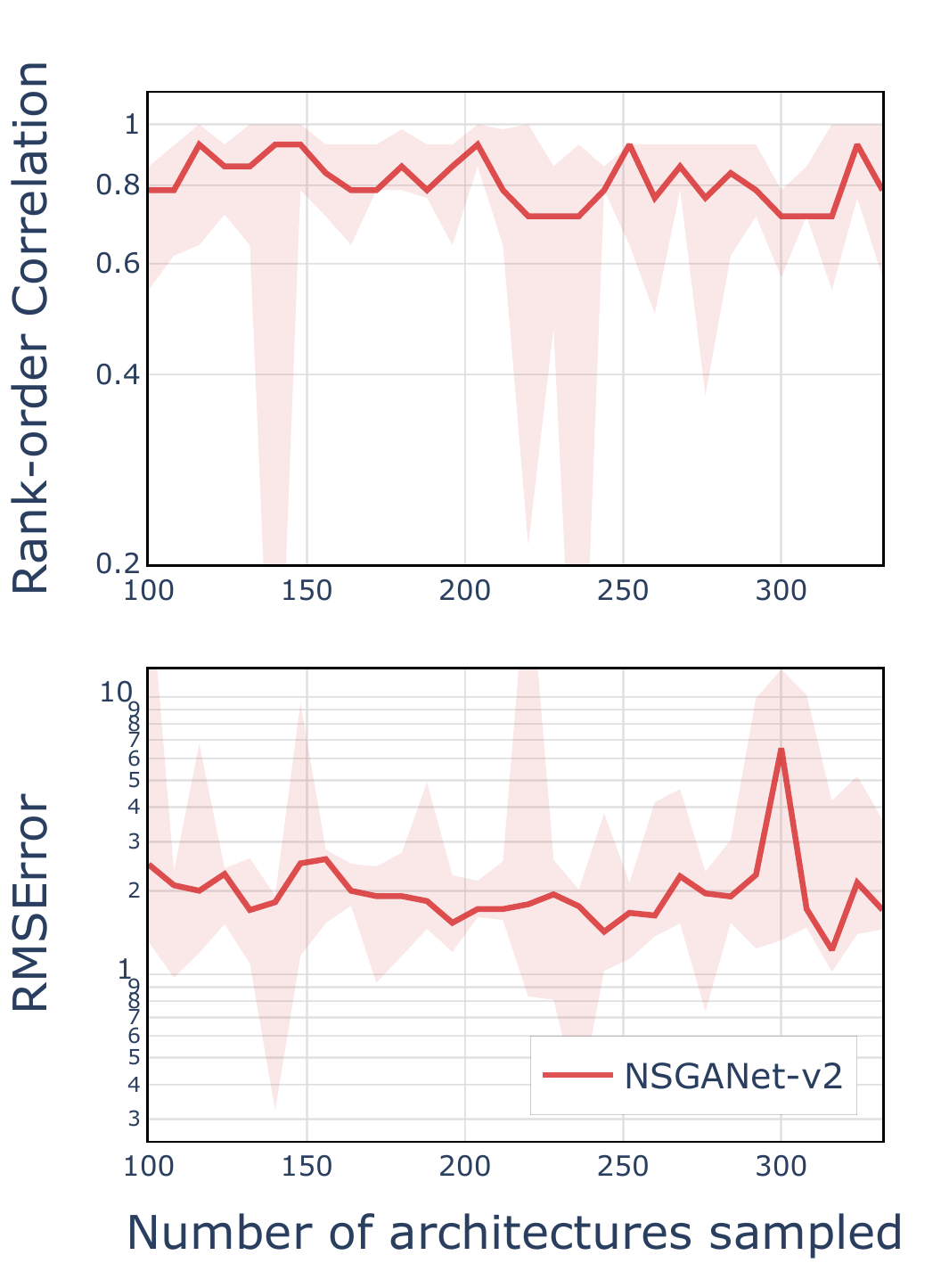}
    \caption{CIFAR-10
    \label{fig:cifar10_search}}
    \end{subfigure}
\caption{
Comparing the relative search efficiency of \ourmethod{} to other methods under bi-objective setup on ImageNet (a) and CIFAR-10 (b). The left plots in each subfigure compares the hypervolume metric \cite{hypervolume}, where a larger value indicates a better Pareto front achieved. The right plots in each subfigure show the Spearman rank-order correlation (top) and the root mean square error (bottom) of \ourmethod{}. All results are averaged over five runs with standard deviation shown in shaded regions. \label{fig:nsganetv2_search_efficiency}}
\end{figure}

In this section, we first compare the search efficiency of \ourmethod{} to other single-objective methods on both CIFAR-10 and ImageNet. To quantify the speedup, we compare the two governing factors, namely, the total number of architectures evaluated by each method to reach the reported accuracy and the number of epochs undertaken to train each sampled architecture during search. The results are provided in Table~\ref{tab:search_cost_breakdown}. We observe that \ourmethod{} is \textbf{20x faster} than methods that use RL or EA. When compared to PNAS \cite{PNAS}, which also utilizes an accuracy predictor, \ourmethod{} is still at least \textbf{3x faster}.

We then compare the search efficiency of \ourmethod{} to NSGANet \cite{NSGANet} and random search under a bi-objective setup: Top-1 accuracy and \#MAdds. To perform the comparison, we run \ourmethod{} for 30 iterations, leading to 350 architectures evaluated in total. We record the cumulative hypervolume \cite{hypervolume} achieved against the number of architectures evaluated. We repeat this process five times on both ImageNet and CIFAR-10 datasets to capture the variance in performance due to randomness in the search initialization. For a fair comparison to NSGANet, we apply the search code to our search space and record the number of architectures evaluated by NSGANet to reach a similar hypervolume than that achieved by \ourmethod{}. The random search baseline is performed by uniformly sampling from our search space. We plot the mean and the standard deviation of the hypervolume values achieved by each method in Fig.~\ref{fig:nsganetv2_search_efficiency}. Based on the incremental rate of hypervolume metric, we observe that \ourmethod{} is \textbf{2 - 5x faster}, on average, in achieving a better Pareto frontier in terms of number of architectures evaluated.

\subsection{Results on Standard Datasets}
Prior to the search, we train the supernet following the training hyperparameters setting from \cite{onceforall}. For each dataset, we start \ourmethod{} with 100 randomly sampled architectures and run for 30 iterations. In each iteration, we evaluate 8 architectures selected from the candidates recommended by NSGA-II according to the accuracy predictor. For searching on CIFAR-10 and CIFAR-100, we fine tune the weights inherited from the supernet for five epochs then evaluate on 5K held-out validation images from the original training set. For searching on ImageNet, we re-calibrate the running statistics of the BN layers after inheriting the weights from the supernet, and evaluate on 10K held-out validation images from the original training set. At the conclusion of the search, we pick the four architectures from the achieved Pareto front, and further fine-tune for additional 150-300 epochs on the entire training sets. For reference purpose, we name the obtained architectures as \ourmodel{}-s/m/l/xl in ascending \#MAdds order. Architectural details can be found in the Appendix \ref{sec:arch}.

\begin{table}[!t]
\caption{ImageNet Classification \cite{imagenet}: comparing \ourmodel{} with manual and automated design of efficient networks. Models are grouped into sections for better visualization. Our results are underlined and best result in each section is in bold. CPU latency (batchsize=1) is measured on Intel i7-8700K and GPU latency (batchsize=64) is measured on 1080Ti. $^\dagger$ The search cost excludes the supernet training cost. $^\ddagger$ Estimated based on the claim that PNAS is 8x faster than NASNet from \cite{PNAS}.}
\label{tab:imagenet}
\centering
\resizebox{0.98\textwidth}{!}{%
\begin{tabular}{@{\hspace{2mm}}l|cc|cc|cc|cc@{\hspace{2mm}}}
\specialrule{1.5pt}{1pt}{1pt}
Model \hspace{1mm} & \hspace{1mm} Type & \hspace{1mm} \begin{tabular}[c]{@{}c@{}}Search Cost\\ (GPU days)\end{tabular} \hspace{1mm} & \hspace{1mm} \#Params & \hspace{1mm} \#MAdds \hspace{1mm} & \hspace{1mm} \begin{tabular}[c]{@{}c@{}}CPU Lat.\\ (ms)\end{tabular} & \hspace{1mm} \begin{tabular}[c]{@{}c@{}}GPU Lat.\\ (ms)\end{tabular} \hspace{1mm} & \hspace{1mm} \begin{tabular}[c]{@{}c@{}}Top-1 Acc.\\ (\%)\end{tabular} & \hspace{1mm} \begin{tabular}[c]{@{}c@{}}Top-5 Acc.\\ (\%)\end{tabular} \\
\specialrule{1.5pt}{1pt}{1pt}
\textbf{\ourmodel{}-s} & auto & 1$^\dagger$ & {\ul 6.1M} & {\ul 225M} & {\ul 9.1} & {\ul 30} & {\ul \textbf{77.4}} & {\ul \textbf{93.5}} \\
MobileNetV2 \cite{mobilenetv2} & manual & 0 & 3.4M & 300M & \textbf{8.3} & \textbf{23} & 72.0 & 91.0 \\
FBNet-C \cite{fbnet} & auto & 9 & 5.5M & 375M & 9.1 & 31 & 74.9 & - \\
ProxylessNAS \cite{proxylessnas} & auto & 8.3 & 7.1M & 465M & 8.5 & 27 & 75.1 & 92.5 \\
MobileNetV3 \cite{mobilenetv3} & combined & - & 5.4M & 219M & 10.0 & 33 & 75.2 & - \\
OnceForAll \cite{onceforall} & auto & 2$^\dagger$ & 6.1M & 230M & 9.5 & 31 & 76.9 & - \\
\midrule
\textbf{\ourmodel{}-m} & auto & 1$^\dagger$ & {\ul 7.7M} & {\ul 312M} & {\ul \textbf{11.4}} & {\ul \textbf{37}} & {\ul \textbf{78.3}} & {\ul \textbf{94.1}} \\
EfficientNet-B0 \cite{efficientnet} & auto & - & 5.3M & 390M & 14.4 & 46 & 76.3 & 93.2 \\
MixNet-M \cite{mixnet} & auto & - & 5.0M & 360M & 24.3 & 79 & 77.0 & 93.3 \\
AtomNAS-C+ \cite{atomnas} & auto & 1$^\dagger$ & 5.5M & 329M & - & - & 77.2 & 93.5 \\ \midrule
\textbf{\ourmodel{}-l} & auto & 1$^\dagger$ & {\ul 8.0M} & {\ul 400M} & {\ul \textbf{12.9}} & {\ul \textbf{52}} & {\ul \textbf{79.1}} & {\ul \textbf{94.5}} \\
PNASNet-5 \cite{PNAS} & auto & 250$^\ddagger$ & 5.1M & 588M & 35.6 & 82 & 74.2 & 91.9 \\ \midrule
\textbf{\ourmodel{}-xl} & auto & 1$^\dagger$ & {\ul 8.7M} & {\ul 593M} & {\ul \textbf{16.7}} & {\ul \textbf{73}} & {\ul \textbf{80.4}} & {\ul \textbf{95.2}} \\
EfficientNet-B1 \cite{efficientnet} & auto & - & 7.8M & 700M & 21.5 & 78 & 78.8 & 94.4 \\
MixNet-L \cite{mixnet} & auto & - & 7.3M & 565M & 29.4 & 105 & 78.9 & 94.2 \\
\specialrule{1.5pt}{1pt}{1pt}
\end{tabular}}
\end{table}

\begin{figure}[t]
    \centering
    \includegraphics[width=0.95\textwidth{}]{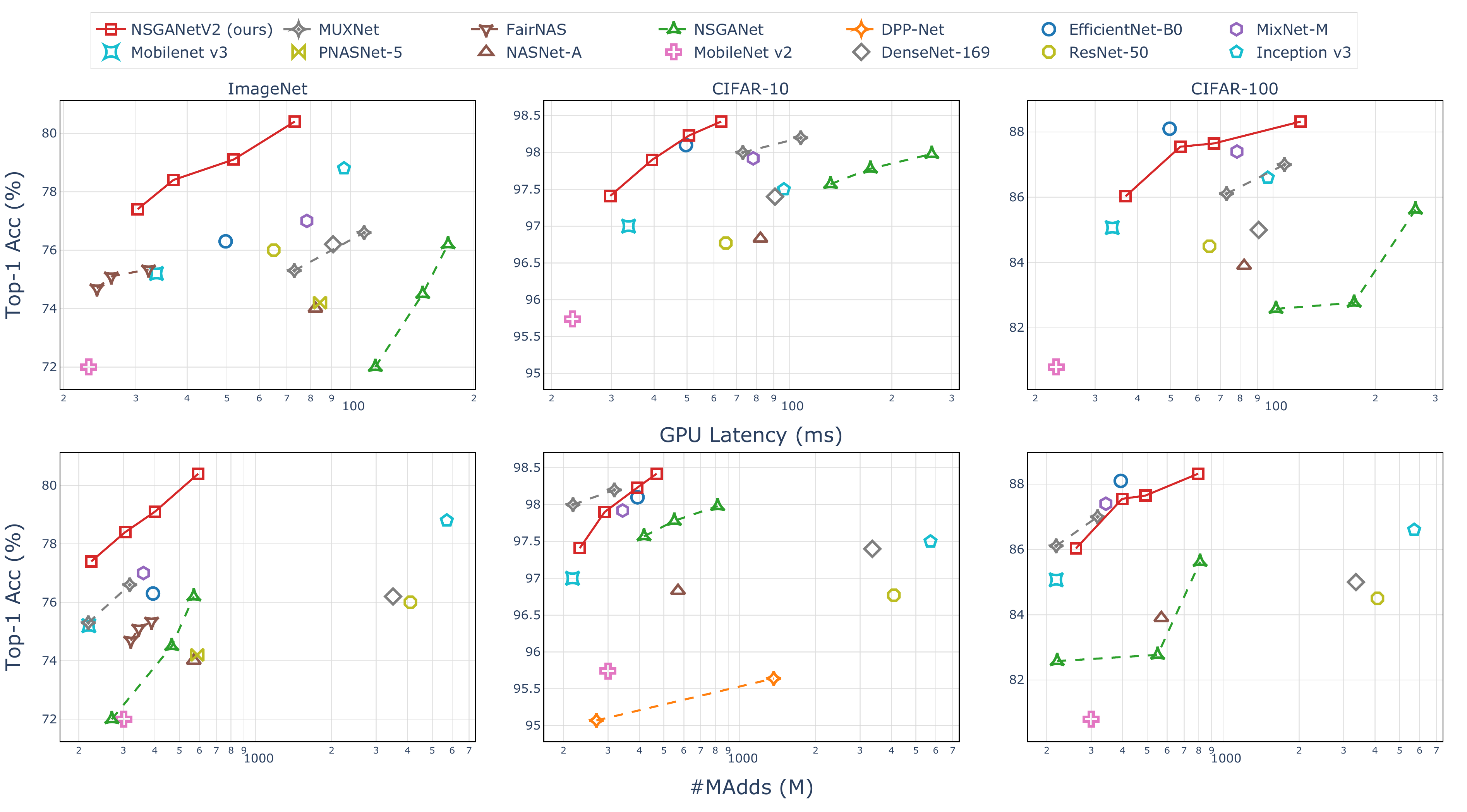}
    \caption{\textbf{Accuracy} vs \textbf{Efficiency}: Top row compares predictive accuracy vs. GPU latency on a batch of 64 images. Bottom row compares predictive accuracy vs. number of multi-adds in millions. Models from multi-objective approaches are joined with lines. Our models are obtained by directly searching on the respective datasets. In most problems, \ourmethod{} finds more accurate solutions with fewer parameters.
    \label{fig:nsganetv2_main}}
\end{figure}

Table~\ref{tab:imagenet} shows the performance of our models on the ImageNet 2012 benchmark \cite{imagenet}. We compare models in terms of predictive performance on the validation set, model efficiency (measured by \#MAdds and latencies on different hardware), and associated search cost. Overall, \ourmodel{} consistently either matches or outperforms other models across different accuracy levels with highly competitive search costs. In particular, \ourmodel{}-s is \textbf{2.2\% more accurate} than MobileNetV3 \cite{mobilenetv3} while being equivalent in \#MAdds and latencies; \ourmodel{}-xl achieves \textbf{80.4\% Top-1 accuracy} under 600M MAdds, which is \textbf{1.5\% more accurate} and \textbf{1.2x more efficient} than EfficientNet-B1 \cite{efficientnet}. Additional comparisons to models from multi-objective approaches are provided in Fig.~\ref{fig:nsganetv2_main}.

For CIFAR datasets, Fig.~\ref{fig:nsganetv2_main} compares our models with other approaches in terms of both predictive performance and computational efficiency. On CIFAR-10, we observe that \ourmodel{} dominates all previous models including (1) NASNet-A \cite{nasnet}, PNASNet-5 \cite{PNAS} and NSGANet \cite{NSGANet} that search on CIFAR-10 directly, and (2) EfficientNet \cite{efficientnet}, MobileNetV3 \cite{mobilenetv3} and MixNet \cite{mixnet} that fine-tune from ImageNet.

\section{Scalability of \ourmethod{}}

\subsection{Types of Datasets}
Existing NAS approaches are rarely evaluated for their search ability beyond standard benchmark datasets, i.e., ImageNet, CIFAR-10, and CIFAR-100. Instead, they follow a conventional transfer learning setup, in which the architectures found by searching on standard benchmark datasets are transferred, with weights fine-tuned, to new datasets. We argue that such a process is conceptually contradictory to the goal of NAS, and the architectures identified under such a process are sub-optimal. In this section we demonstrate the scalability of \ourmethod{} to six additional datasets with various forms of difficulties, in terms of diversity in classification classes (multi-classes vs. fine-grained) and size of training set (see Table~\ref{tab:dataset}). We adopt the settings of the CIFAR datasets as outlined in Section~\ref{sec:approach}. For each dataset, one search takes less than one day on 8 GPU cards. \
\begin{wraptable}{r}{8cm}
\centering
\resizebox{0.60\textwidth}{!}{%
\begin{tabular}{|l|c|c|cc|}
\hline
Datasets \hspace{1mm} &\hspace{1mm} Type \hspace{1mm} & \hspace{1mm} \#Classes \hspace{1mm} & \hspace{1mm} \#Train & \hspace{1mm} \#Test \\ \hline
CINIC-10 \cite{cinic10} & multi-class & 10 & 90,000 & 90,000 \\
STL-10 \cite{stl-10} & multi-class & 10 & 5,000 & 8,000 \\
Flowers102 \cite{flowers102} & fine-grained & 102 & 2,040 & 6,149 \\
Pets \cite{pets} & fine-grained & 37 & 3,680 & 3,369 \\
DTD \cite{dtd} & fine-grained & 47 & 3,760 & 1,880 \\
Aircraft \cite{aircraft} & fine-grained & 100 & 6,667 & 3,333 \\ \hline
\end{tabular}%
}
\caption{Non-standard Datasets for \ourmethod{}
\label{tab:dataset}}
\label{table:ta2}
\end{wraptable}

\begin{figure}[t]
    \centering
    \begin{subfigure}{0.95\textwidth}
    \centering
    \includegraphics[width=\textwidth{}]{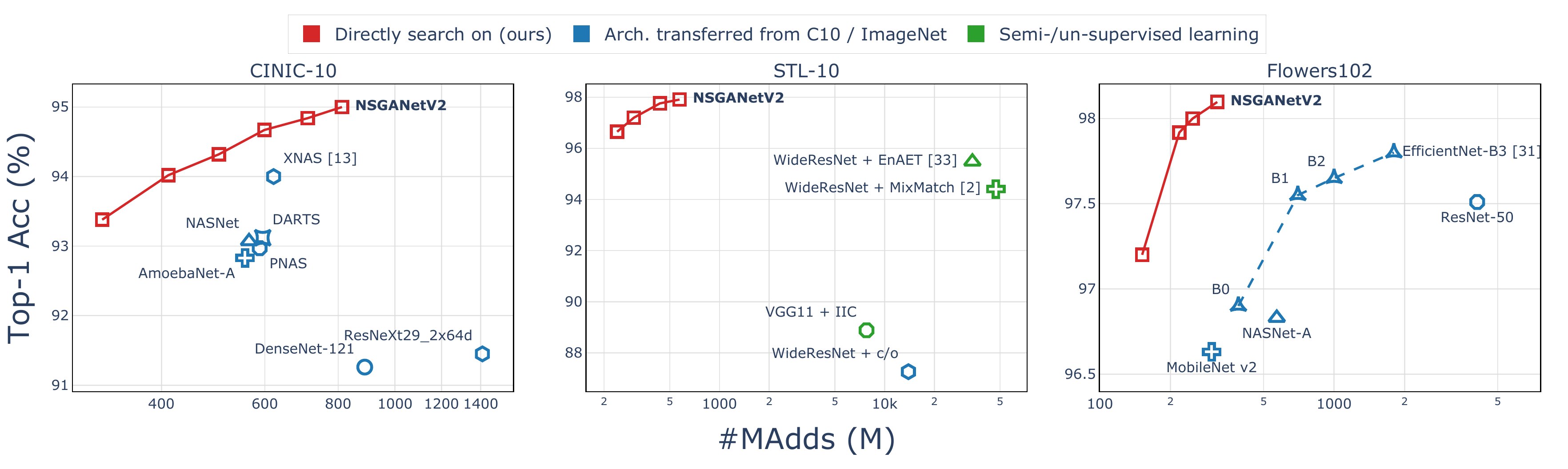}
    \end{subfigure}
    \begin{subfigure}{0.95\textwidth}
    \centering
    \includegraphics[width=\textwidth{}]{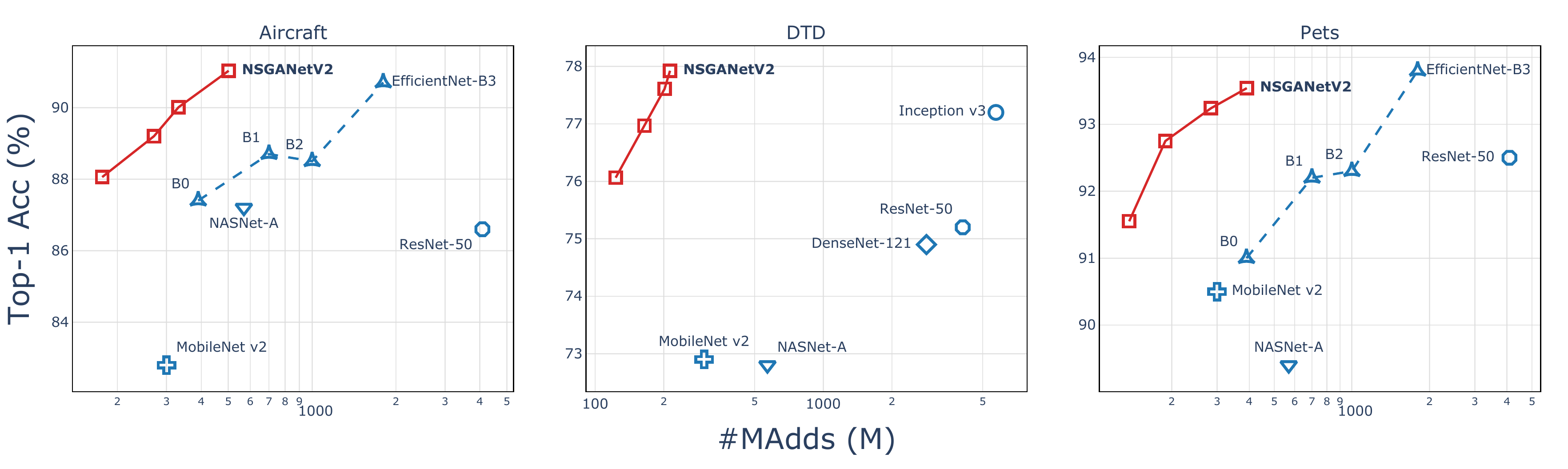}
    \end{subfigure}
\caption{Performance of the set of task-specific models, i.e. \ourmodel{}s, on six different types of non-standard datasets, compared to SOTA from transfer learning \cite{xnas,efficientnet} and semi-/un-supervised learning \cite{mixmatch,wang2019enaet}.\label{fig:non-standard-datasets}}
\end{figure}

Fig.~\ref{fig:non-standard-datasets} (Bottom) compares the performance of \ourmodel{} obtained by searching directly on the respective datasets to models from other approaches that transfer architectures learned from either CIFAR-10 or ImageNet. Overall, we observe that \ourmodel{} significantly outperforms other models on all three datasets. In particular, \ourmodel{} achieves a better performance than the currently known state-of-the-art on CINIC-10 \cite{xnas} and STL-10 \cite{mixmatch}. Furthermore, on Oxford Flowers102, \ourmodel{} achieves better accuracy to that of EfficientNet-B3 \cite{efficientnet} while using \textbf{1.4B fewer} MAdds.

\subsection{Number of Objectives}
\begin{figure}[!tbh]
    \begin{subfigure}[b]{0.38\textwidth}
    \centering
    \includegraphics[width=\textwidth{}]{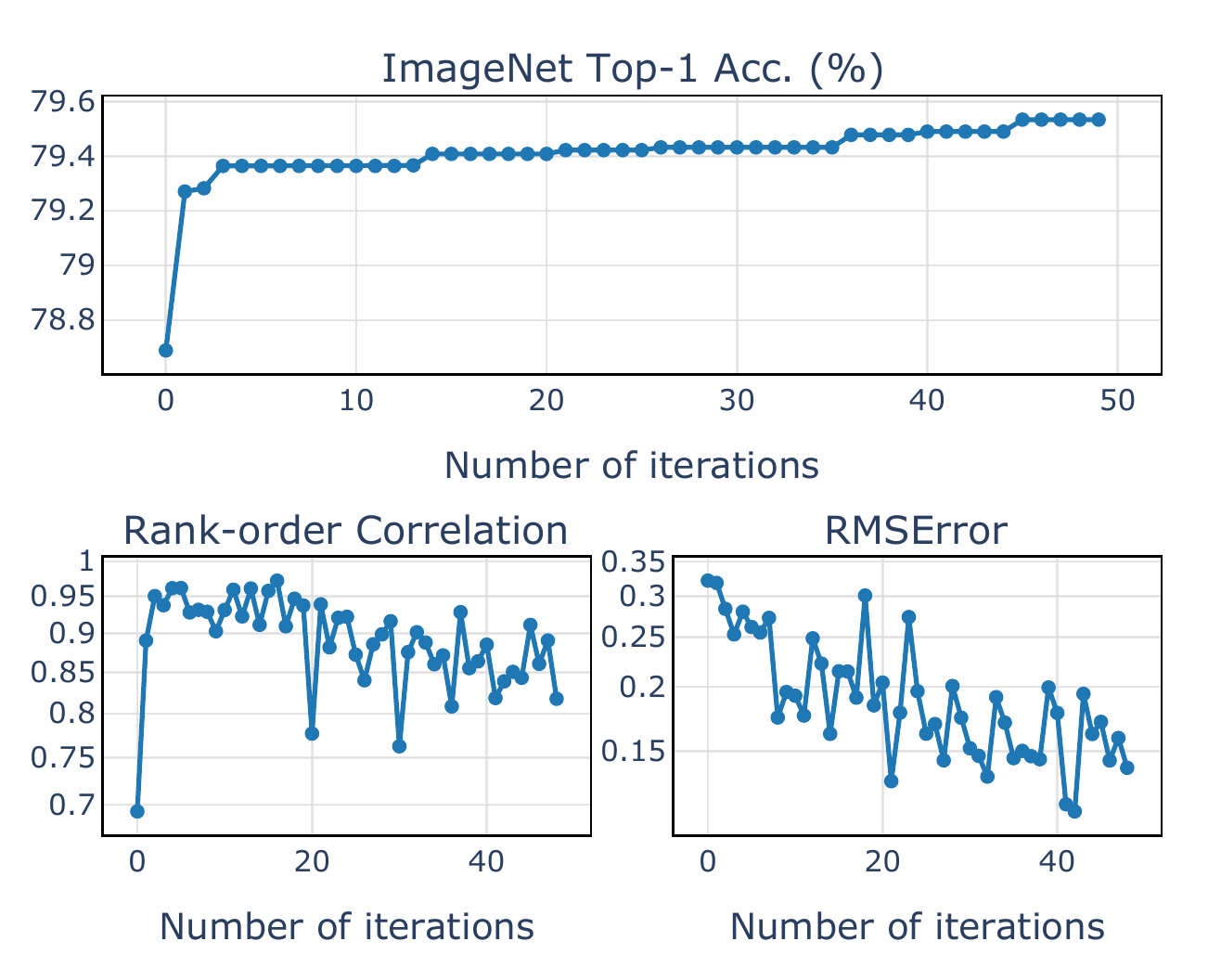}
    \caption{Maximize Top-1 Acc.
    \label{fig:nsganetv2_single}}
    \end{subfigure}\hfill
    \begin{subfigure}[b]{0.6\textwidth}
    \centering
    \includegraphics[width=\textwidth{}]{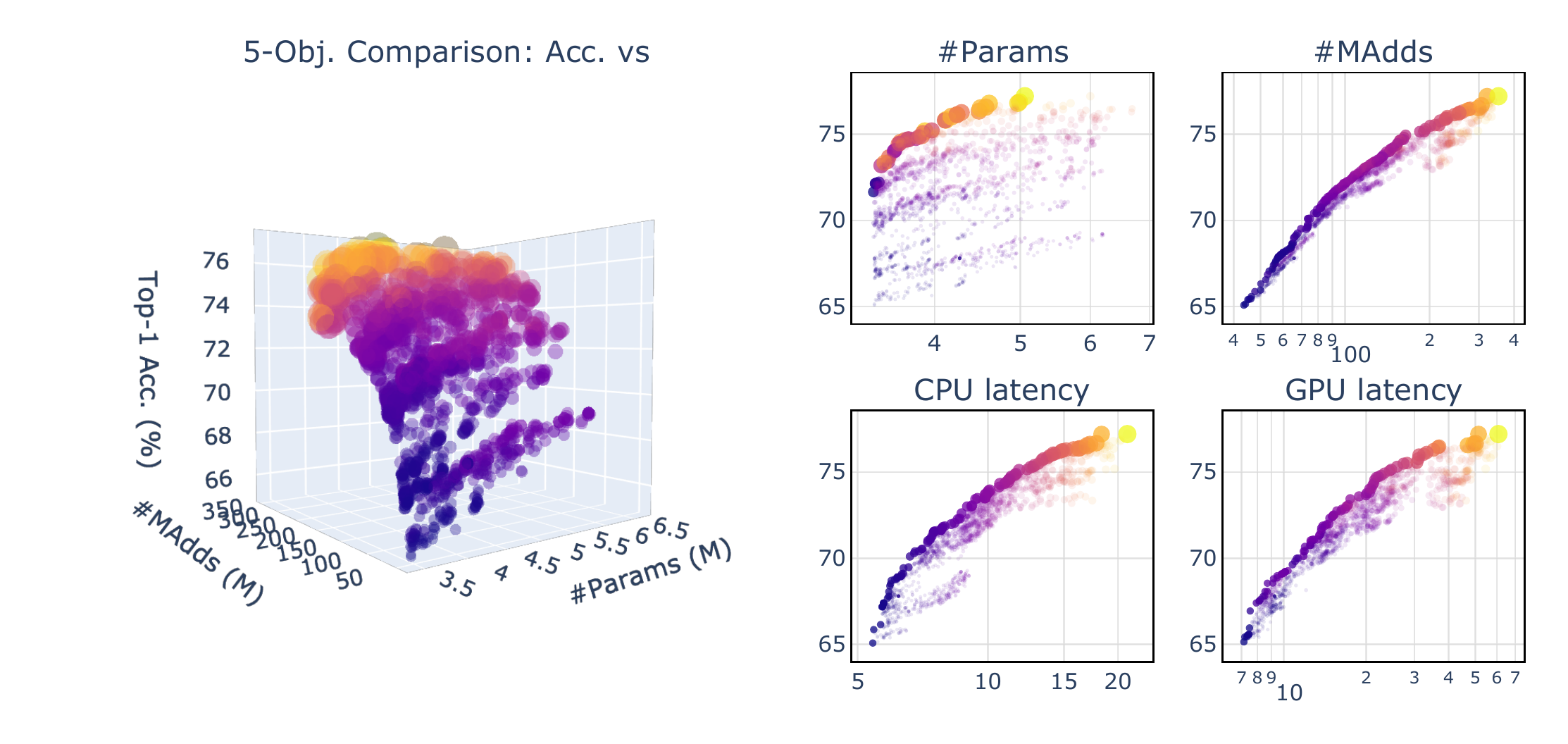}
    \caption{5-Objective scenario.
    \label{fig:nsganetv2_5obj}}
    \end{subfigure}\\
    \centering
    \begin{subfigure}{0.95\textwidth}
    \includegraphics[width=\textwidth{}]{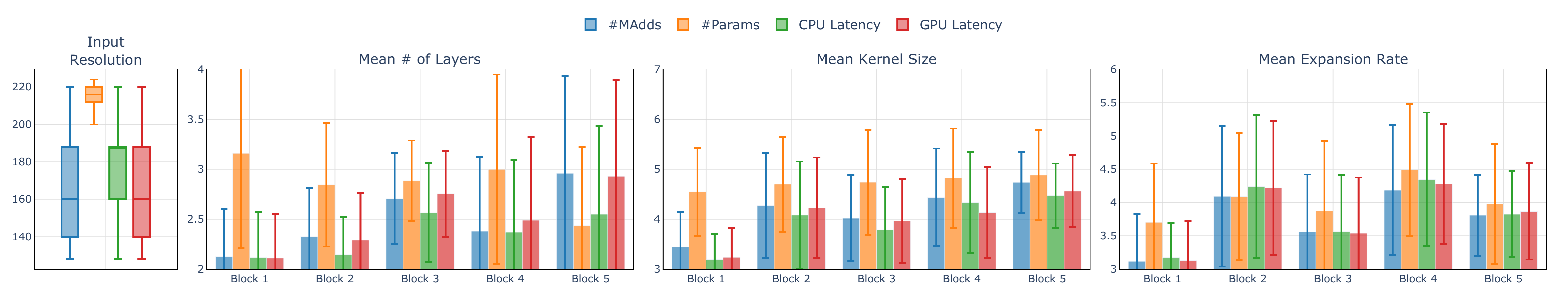}
    \caption{Non-dominated architectures under different efficiency objectives.
    \label{fig:nsganetv2_innovization}}
    \end{subfigure}
\caption{Scalability of \ourmethod{} to different numbers and types of objectives: optimizing (a) a scalarized single-objective on ImageNet; (b) five objectives including accuracy, Params, MAdds, CPU and GPU latency, simultaneously. (c) Post-optimal analysis on the architectures that are non-dominated according to different efficiency objectives.}
\label{fig:image2}
\end{figure}

\noindent\textbf{Single-objective Formulation:} Adding a hardware efficiency target as a penalty term to the objective of maximizing predictive performance is a common workaround to handle multiple objectives in the NAS literature \cite{proxylessnas,mnasnet,fbnet}. We demonstrate that our proposed algorithm can also effectively handle such a scalarized single-objective search. Following the scalarization method in \cite{mnasnet}, we apply \ourmethod{} to maximize validation accuracy on ImageNet with 600M MAdds as the targeted efficiency. The accumulative top-1 accuracy achieved and the performance of the accuracy predictor are provided in Fig.~\ref{fig:nsganetv2_single}. Without further fine-tuning, the obtained architecture yields 79.56\% accuracy with 596M MAdds on the ImageNet validation set, which is more accurate and \textbf{100M fewer} MAdds than EfficientNet-B1 \cite{efficientnet}.

\noindent\textbf{Many-objective Formulation:} Practical deployment of learned models are rarely driven by a single objective, and most often, seek to trade-off many different, possibly competing, objectives. As an example of one such scenario, we use \ourmethod{} to simultaneously optimize five objectives---namely, the accuracy on ImageNet, \#Params, \#MAdds, CPU and GPU latency. We follow the same search setup as in the main experiments and increase the budget to ensure a thorough search on the expanded objective space. We show the obtained Pareto-optimal (to five objectives) architectures in Fig.~\ref{fig:nsganetv2_5obj}. We use color and marker size to indicate CPU and GPU latency, respectively. We observe that a Pareto surface emerges, shown in the left 3D scatter plot, suggesting that trade-offs exist between objectives, i.e., \#Params and \#MAdds are not fully correlated. We then project all architectures to 2D, visualizing accuracy vs. each one of the four considered efficiency measurements, and highlight the architectures that are non-dominated in the corresponding two-objective cases. We observe that many architectures that are non-dominated in the five-objective case are now dominated when only considering two objectives. Empirically, we observe that accuracy is highly correlated with \#MAdds, CPU and GPU latency, but not with \#Params, to some extent.

\section{Conclusion}
This paper introduced \ourmethod{}, an efficient neural architecture search algorithm for rapidly designing task-specific models under multiple competing objectives. The efficiency of our approach stems from (i) online surrogate-modeling at the level of the architecture to improve the sample efficiency of search, and (ii) a supernet based surrogate-model to improve the weights learning efficiency via fine-tuning. On standard datasets (CIFAR-10, CIFAR-100 and ImageNet), \ourmodel{} matches the state-of-the-art with a search cost of one day. The utility and versatility of \ourmethod{} are further demonstrated on non-standard datasets of various types of difficulties and on different number of objectives. Improvements beyond the state-on-the-art on STL-10 and Flowers102 (under mobile setting) suggest that NAS is a more effective alternative to conventional transfer learning approaches.

\bibliographystyle{splncs04}
\bibliography{egbib}


\appendix
\section*{Appendix}
Recall that Neural Architecture Search (NAS) is formulated as a bi-level optimization problem in the original paper. The key idea of \ourmethod{} is to adopt a surrogate model at both the upper and lower level in order to improve the efficiency of solving the NAS bi-level problem. In this appendix, we include the following material:
\begin{enumerate}
\item Further analysis on the upper level surrogate model of \ourmethod{} in Section~\ref{sec:surrogate}.
\item Post-search analysis in terms of mining for \textbf{\emph{architectural design insights}} in Section~\ref{sec:insights}.
\item \textbf{\emph{``Objective transfer"}} in Section~\ref{sec:transfer}. Here we seek to quickly search for architectures optimized for target objectives by initializing the search with architectures sampled from \emph{insights} gained by searching on source objectives.
\item The visualization of the final architectures on the six datasets that we searched in Section~\ref{sec:arch}.
\end{enumerate}

\section{Correlation Between Search Performance and Surrogate Model\label{sec:surrogate}}
In \ourmethod{}, we use a surrogate model at the upper architecture level to reduce the number of architectures sent to the lower level for weight learning. There are at least two desired properties of a surrogate model, namely:

\begin{itemize}
    \item a high rank-order correlation between the performance predicted by the surrogate model and the true performance
    \item a high sample-efficiency such that the number of architectures, that are fully trained and evaluated, for constructing the surrogate model is as low as possible
\end{itemize}

\begin{figure}[!bht]
    \centering
    \includegraphics[width=0.95\textwidth{}]{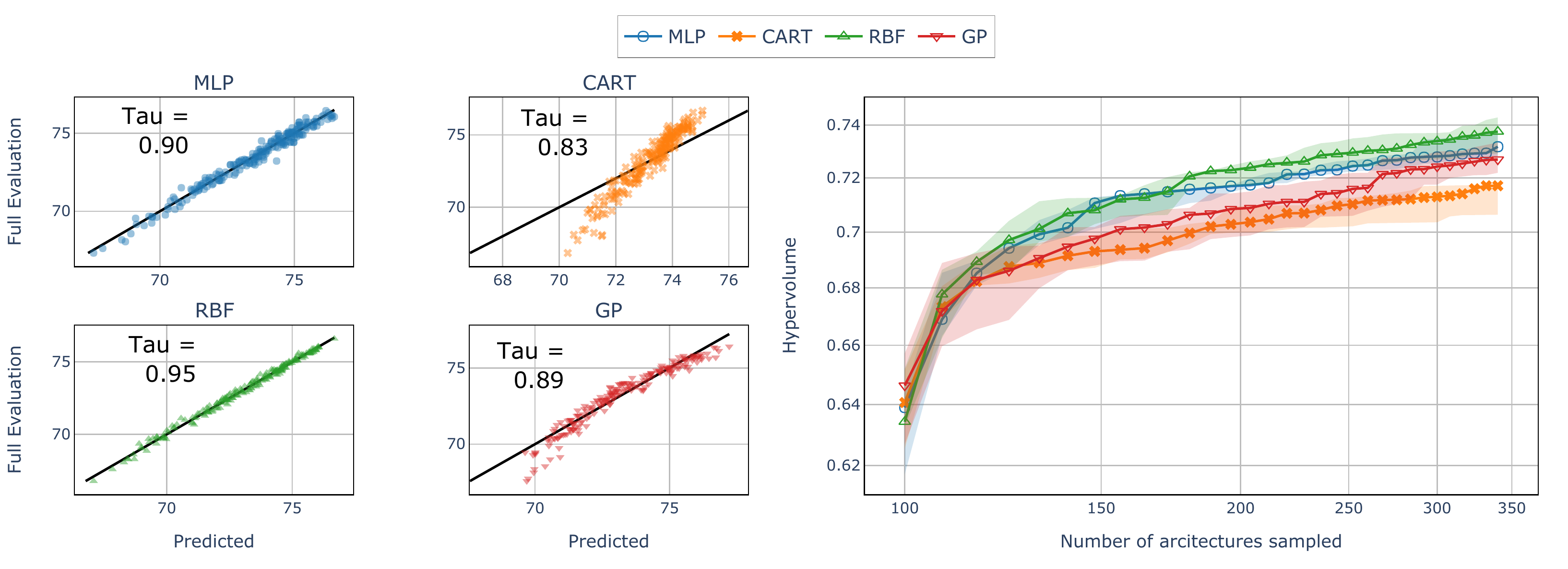}
    \caption{Left: Kendall’s Tau \cite{kendalltau} rank-order correlation comparison among different surrogate models. For each model, we calculate the correlation on 350 architectures fully trained and evaluated during the search and we report the mean value over five runs. Right: \ourmethod{} search performance, measured by hypervolume \cite{hypervolume}, with different surrogate models. Empirically, we observe a positive correlation between rank-order correlation in surrogate model predictions and the search performance when paired with \ourmethod{}. All experiments are performed on ImageNet \cite{imagenet}. In general, the rank-order correlation of \ourmethod{} $(\sim 0.9)$ is significantly better than that achieved by ProgressiveNAS~\cite{PNAS} (0.476).
    \label{fig:rank_order_performance}}
\end{figure}

In this section, we aim to quantify the correlation between the surrogate model's rank-order correlation (Kendall' Tau \cite{kendalltau}) and \ourmethod{}'s search performance. On ImageNet dataset, we run \ourmethod{} with four different surrogate models, including Multi-Layer Perceptron (MLP), Classification And Regression Trees (CART), Radial Basis Function (RBF) and Gaussian Processes (GP). We record the accumulative hypervolume \cite{hypervolume} and calculate the rank-order correlation on all architectures evaluated during the search. The results are provided in Fig.~\ref{fig:rank_order_performance}. In \ourmethod{}, we iteratively fit and refine surrogate models using only architectures that are close to the Pareto frontier. Hence, surrogate models can focus on interpolating across a much restricted region (models close to the current pareto front) in the search space, leading to a significant better rank-order correlation achieved as opposed to existing methods \cite{PNAS,dppnet}, i.e., $\sim 0.9$ for \ourmethod{} vs 0.476 for ProgressiveNas~\cite{PNAS}. Furthermore, we empirically observe that high rank-order correlation in a surrogate model translates into better search performance (lower sample complexity), measured by hypervolume \cite{hypervolume}, when paired with \ourmethod{}. On ImageNet, RBF outperforms the other three  surrogate models considered. However, to improve generalization to other datasets, we follow an adaptive switching routine that compares all four surrogate models and selects the best based on cross validation (see Section 3.3 in the main paper).

\section{Post Search Analysis\label{sec:post}}
\subsection{Mining for Insights\label{sec:insights}}
\begin{figure}[h]
	\centering
	\begin{subfigure}[b]{.48\textwidth}
		\centering
		\includegraphics[width=0.98\textwidth]{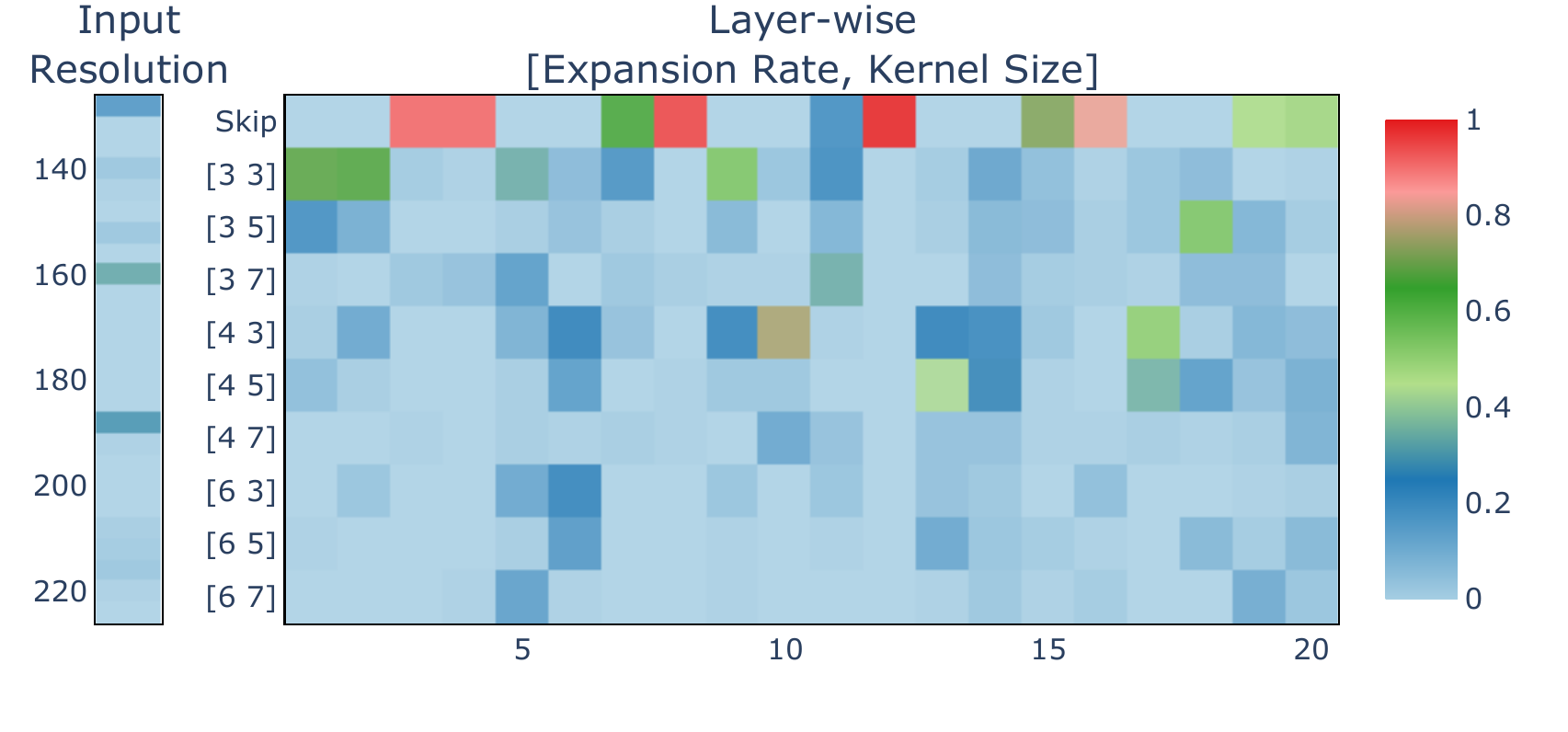}
		\caption{MAdds\label{fig:layer_wise_flops}}
	\end{subfigure} \hfill
	\begin{subfigure}[b]{.48\textwidth}
		\centering
		\includegraphics[width=0.98\textwidth]{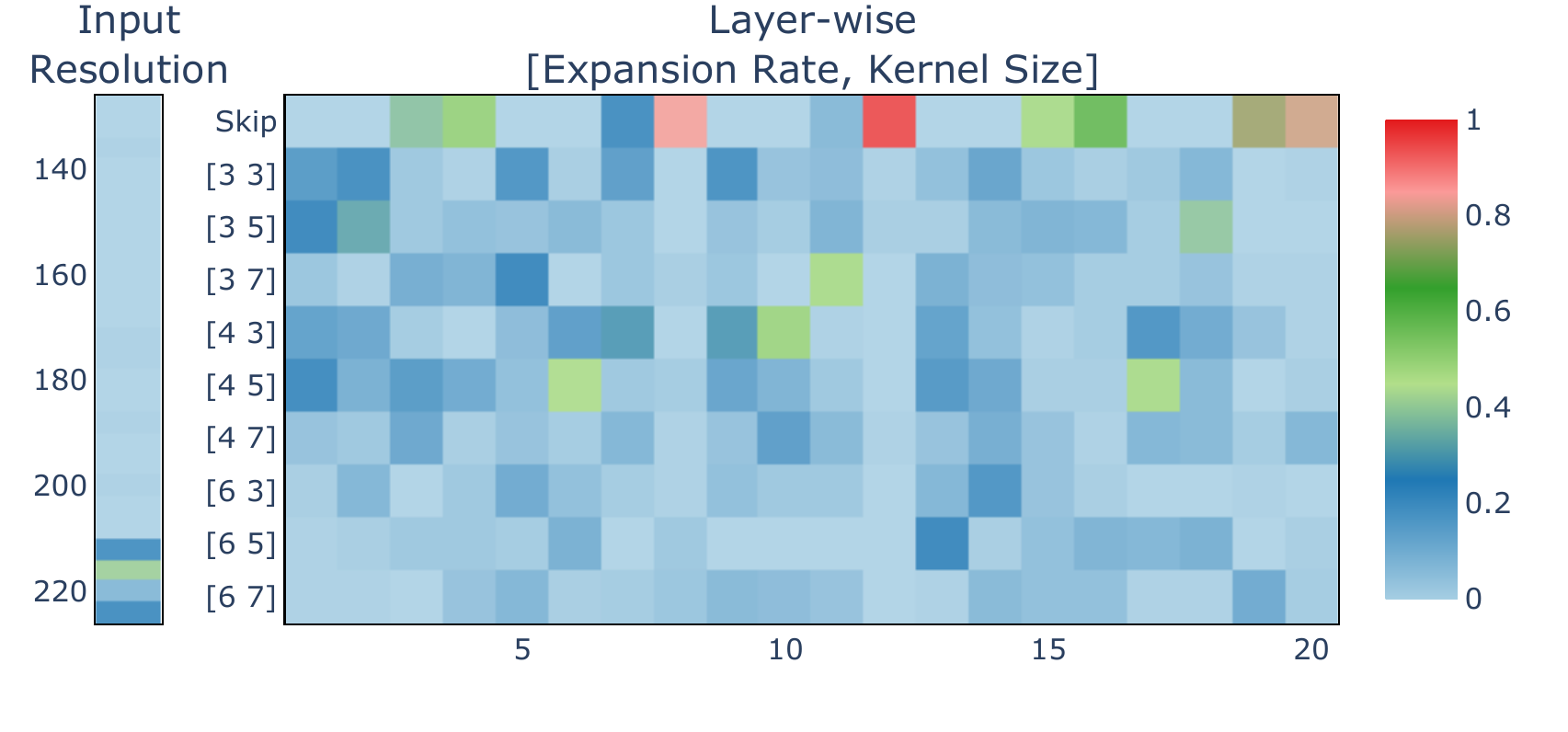}
		\caption{Params\label{fig:layer_wise_params}}
	\end{subfigure} \\ \vspace{2mm}
	\begin{subfigure}[b]{.48\textwidth}
		\centering
		\includegraphics[width=0.98\textwidth]{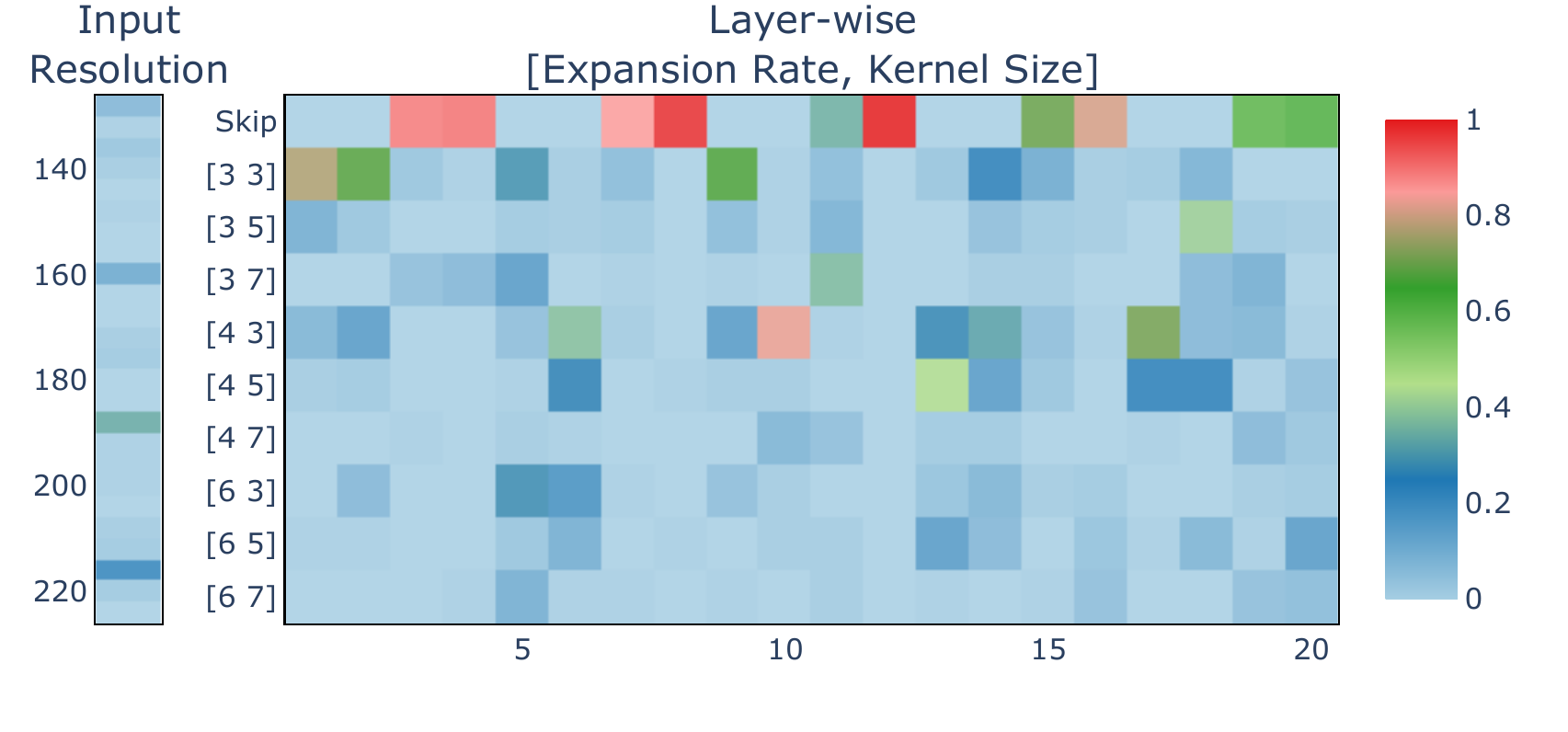}
		\caption{CPU latency\label{fig:layer_wise_cpu}}
	\end{subfigure} \hfill
	\begin{subfigure}[b]{.48\textwidth}
		\centering
		\includegraphics[width=0.98\textwidth]{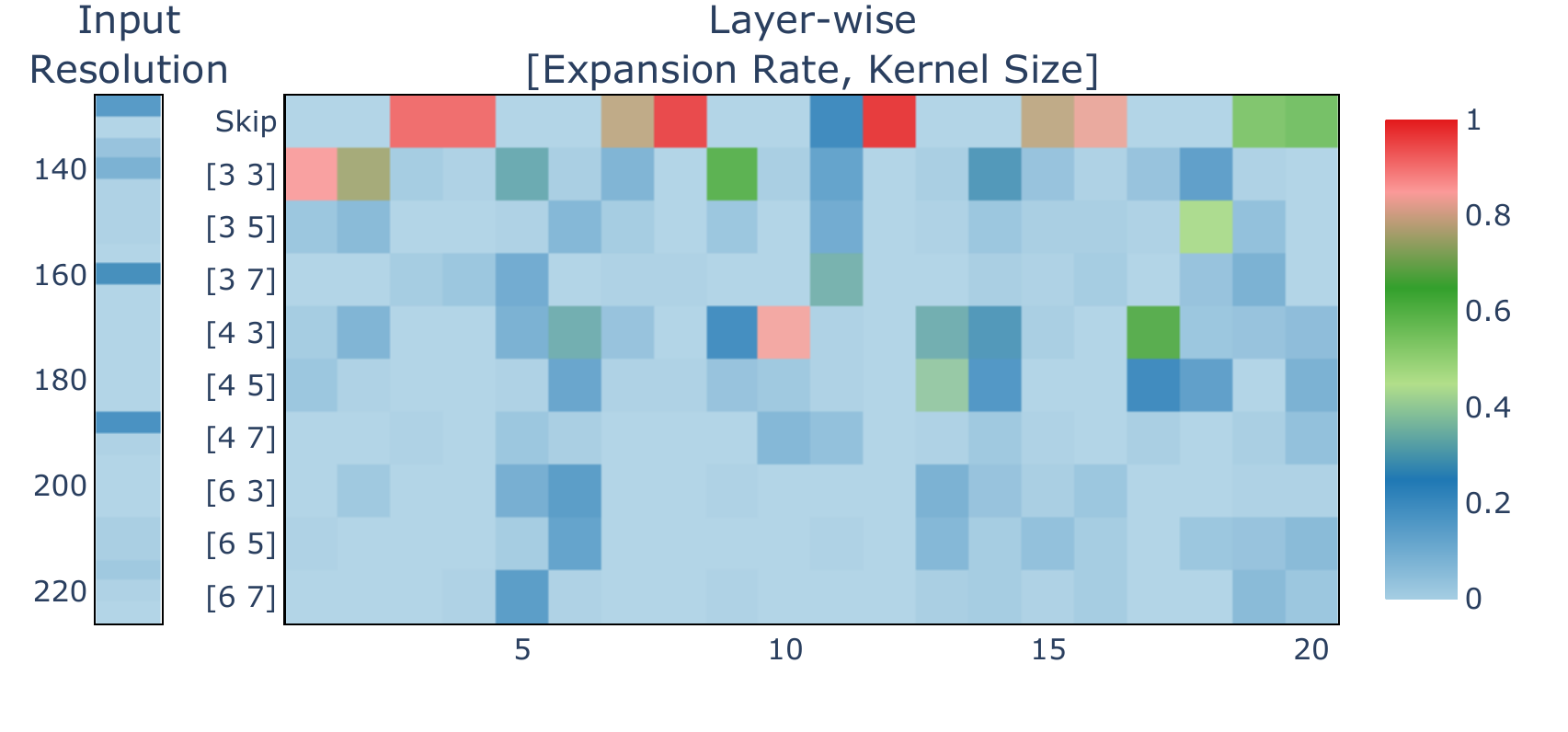}
		\caption{GPU latency\label{fig:layer_wise_gpu}}
	\end{subfigure}
	\caption{The layer-wise architectural choice frequency of the non-dominated architectures obtained by \ourmethod{} when optimizing predictive performance and MAdds (a) / Params (b) / CPU (c) / GPU latency (d).
	\label{fig:layer_wise_distribution}}
\end{figure}
Every single run of \ourmethod{} generates a set of architectures. Mining the information that is generated through that process allows practitioners to choose a suitable architecture \emph{a posteriori} to the search. To demonstrate one such scenario, we ran \ourmethod{} to optimize the predictive performance along with one of  four different efficiency related measurements, namely MAdds, Params, CPU and GPU latency. At the end of the evolution, we identify the non-dominated architectures and visualize their architectural choices in Fig.~\ref{fig:layer_wise_flops} - \ref{fig:layer_wise_gpu}. We observe that the efficient architectures under MAdds, CPU and GPU latency requirements are similar, indicating positive correlation among them, which is not the case with Params. We notice that \ourmethod{} implicitly exploits the fact that Params is agnostic to the image resolution, and choose to use input images at highest allowed resolution to improve predictive performance (see the Input Resolution heatmap in Fig.~\ref{fig:layer_wise_params}).

\subsection{Transfer Across Objectives\label{sec:transfer}}
\begin{figure}[!bht]
    \centering
    \includegraphics[width=0.8\textwidth{}]{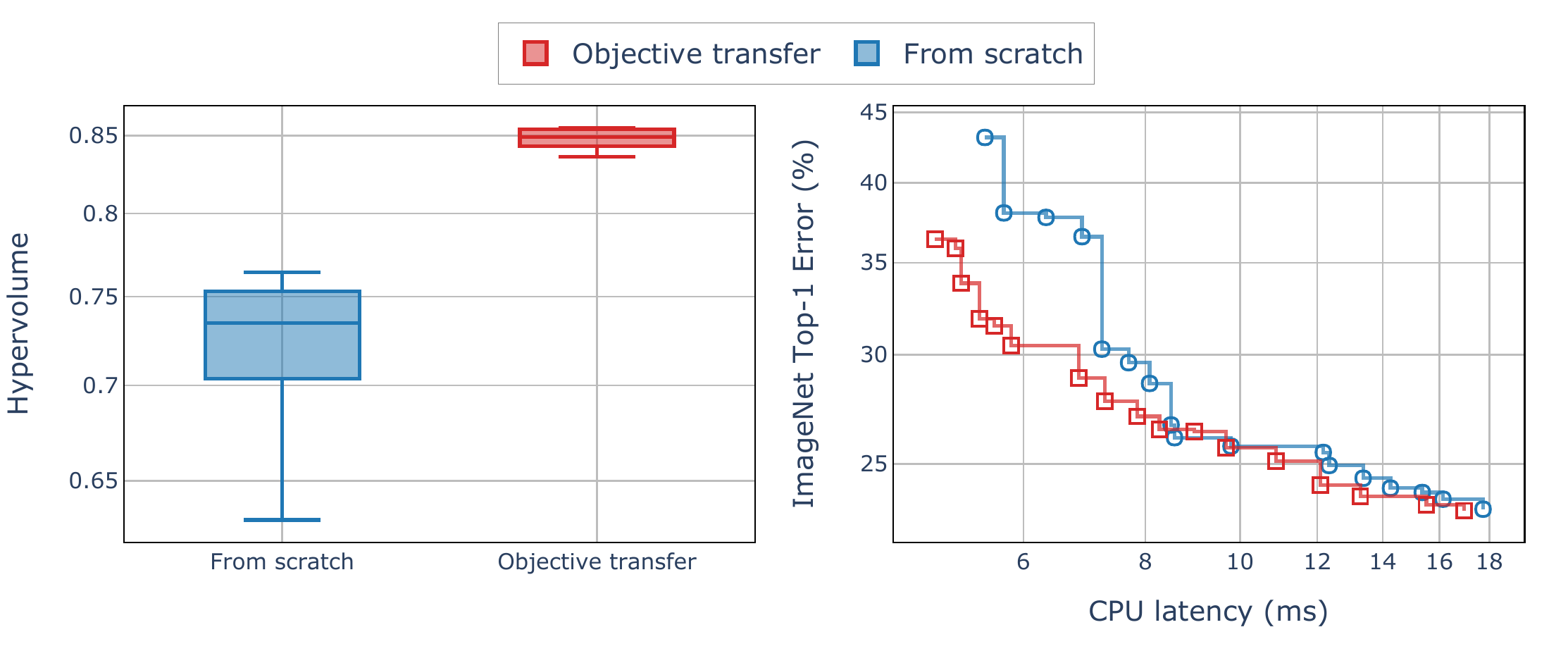}
    \caption{Comparing \ourmethod{}'s search performance when initialized (1) from randomly sampled architectures and (2) by sampling from distribution constructed from efficient architectures on a related task (MAdds).
    \label{fig:innovization}}
\end{figure}
Further post-optimal analysis of the set of non-dominated architectures often times reveals valuable design principles, referred to as derived heuristics \cite{derived_heuristics}. Such derived heuristics can be utilized for novel tasks. Here we consider one such example, transferring  architectures and  associated weights from models that were searched with respect to one pair of objectives to architectures that are optimal with respect to a different pair of objectives. The idea is that if the objectives that we want to transfer across are related but not identical, for instance, MAdds and Latency, we can improve search efficiency by exploiting such correlations. More specifically, we can search for  optimal architectures with respect to a target set of objectives by initializing the search with  architectures from a source set of objectives much more efficiently,  compared to starting the search from scratch. As a demonstration of this property, we conduct the following experiment:
\begin{itemize}
    \item Target Objectives: predictive performance and CPU latency.
    \item Approach 1 (``from scratch''): \ourmethod{} from randomly (uniformly) initialized architectures.
    \item Approach 2 (``from objective transfer''): \ourmethod{} from architectures sampled from distribution constructed from non-dominated architectures of source objectives, namely, predictive performance and MAdds (Fig.~\ref{fig:layer_wise_flops}).
\end{itemize}

In Approach 1, we initialize the search process for the target objectives from randomly sampled architectures (uniformly on the search space). In contrast, in Approach 2, we initialize the search process for the target objectives by architectures sampled from the \emph{insights} obtained by searching on a related pair of source objectives (predictive performance and MAdds) i.e., from the distribution in Fig. \ref{fig:layer_wise_params}(a).

In Fig.~\ref{fig:innovization}, we first compare the  hypervolume achieved by these two approaches over five runs. We visualize the obtained Pareto front (from the run with median hypervolume) in Fig.~\ref{fig:innovization} (Right). We observe that utilizing  \emph{insights} from searching on related objectives can significantly improve  search performance. In general, we believe that heuristics can be derived and utilized to improve search performance on related tasks (e.g. MAdds and CPU latency), which is another desirable property of \ourmethod{}, where a set of architectures are obtained in a single run. The efficiency gains from the objective transfer (Approach 2) we demonstrate here are directly proportional to the correlation between the source and target objectives. However, if the source and target objectives are not related, then Approach 2 may not be more efficient than Approach 1.

\begin{figure}[!h]
    \centering
    \begin{subfigure}{0.65\textwidth}
    \centering
    \includegraphics[width=\textwidth{}]{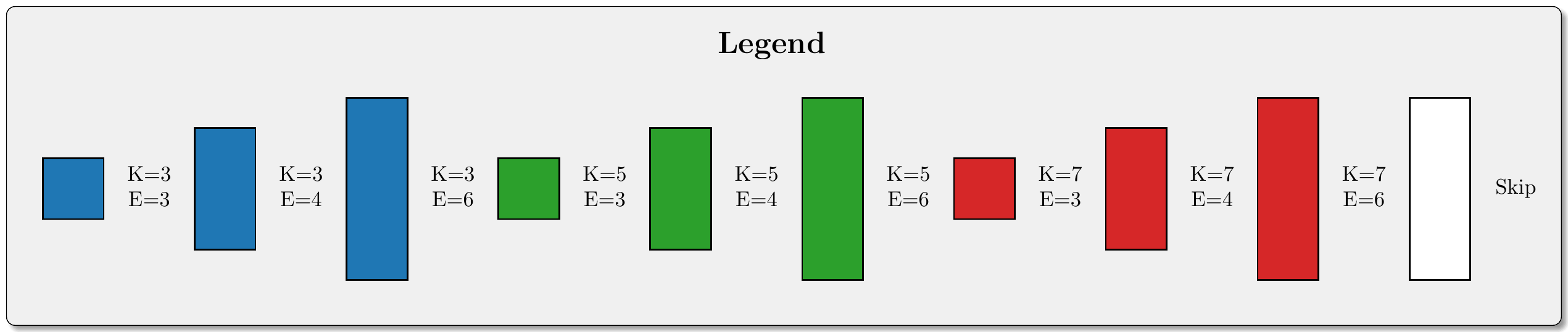}
    \end{subfigure}\\ \vspace{3mm}
    \begin{subfigure}{0.48\textwidth}
    \centering
    \includegraphics[width=\textwidth{}]{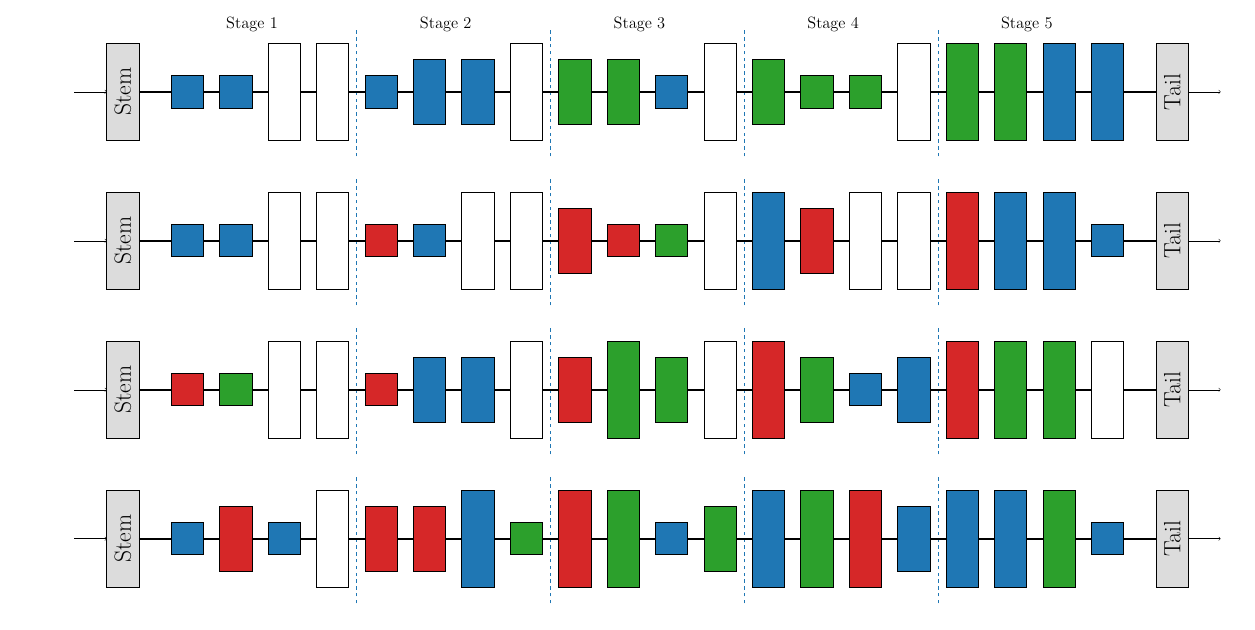}
    \caption{ImageNet}
    \end{subfigure}\hfill
    \begin{subfigure}{0.48\textwidth}
    \centering
    \includegraphics[width=\textwidth{}]{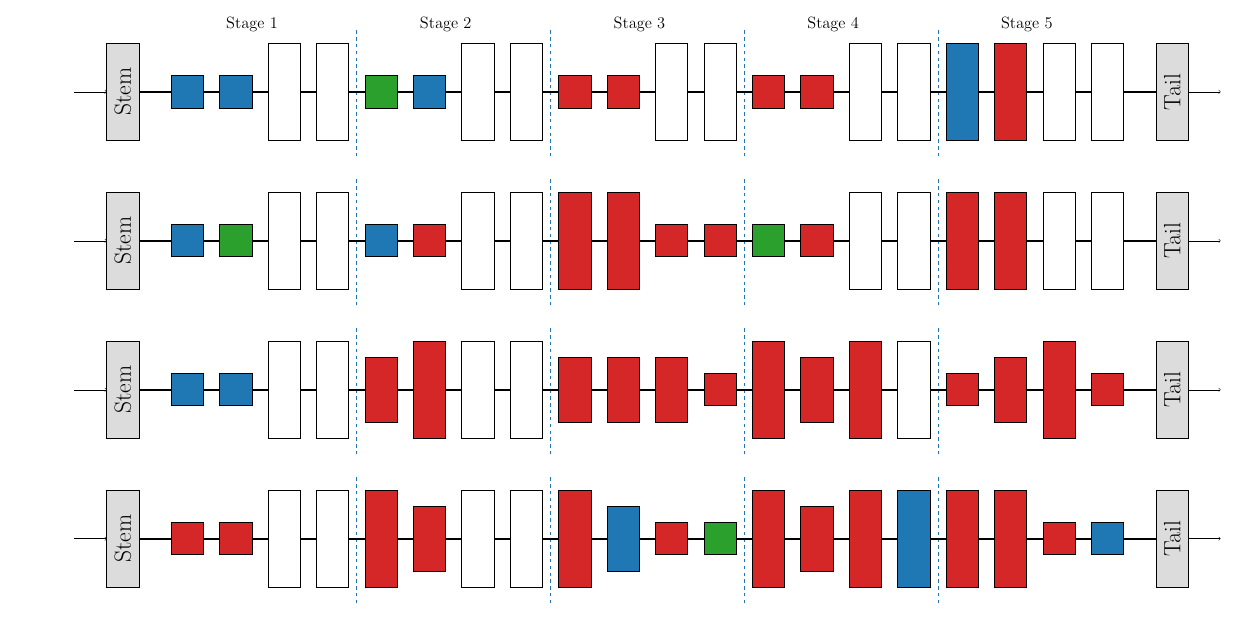}
    \caption{CIFAR-10}
    \end{subfigure}\\
    \begin{subfigure}{0.48\textwidth}
    \centering
    \includegraphics[width=\textwidth{}]{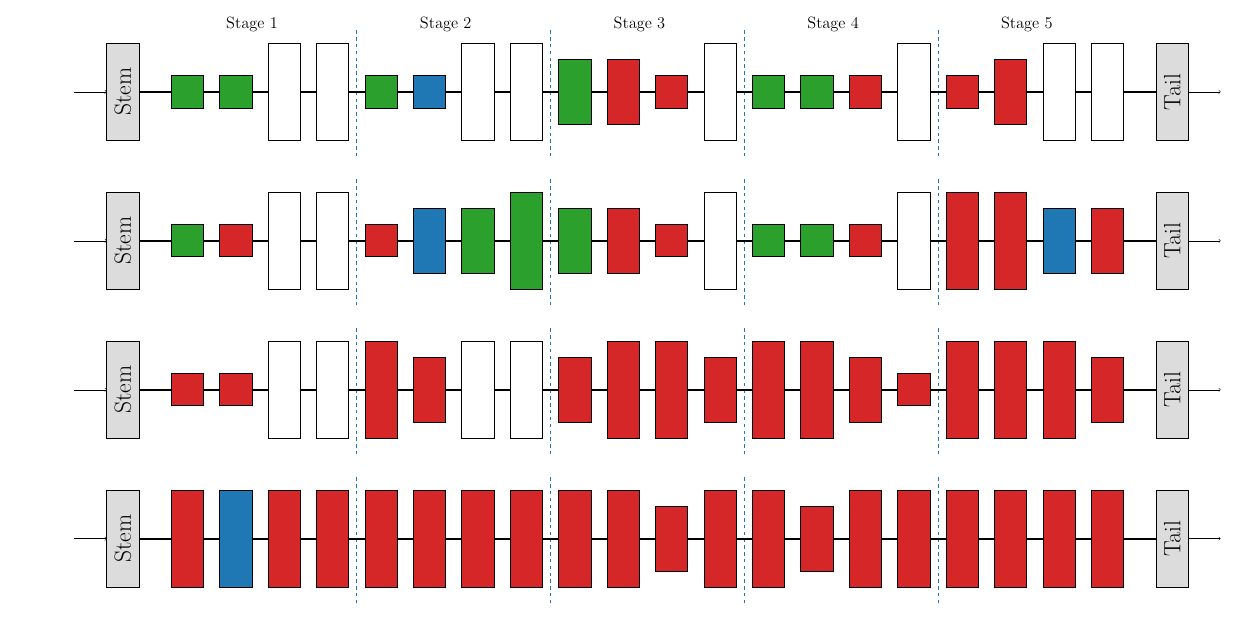}
    \caption{CIFAR-100}
    \end{subfigure}\hfill
    \begin{subfigure}{0.48\textwidth}
    \centering
    \includegraphics[width=\textwidth{}]{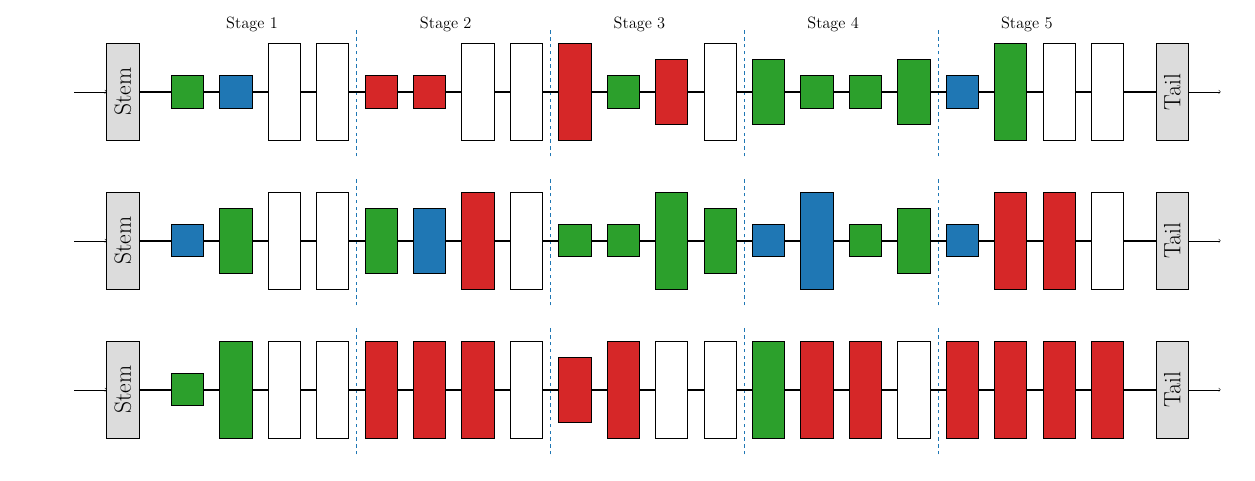}
    \caption{CINIC-10}
    \end{subfigure}\\
    \begin{subfigure}{0.48\textwidth}
    \centering
    \includegraphics[width=\textwidth{}]{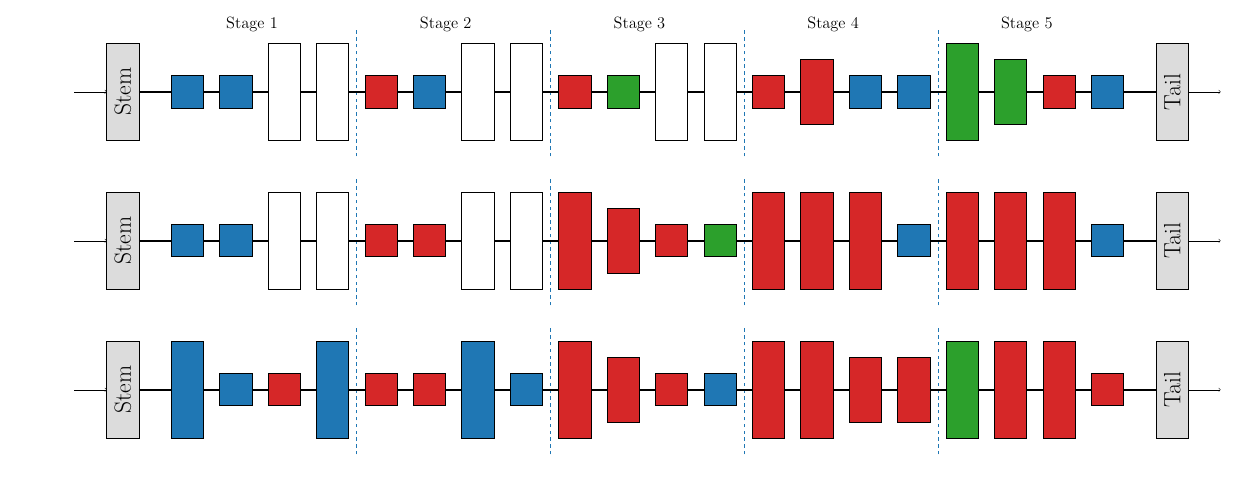}
    \caption{STL-10}
    \end{subfigure}\hfill
    \begin{subfigure}{0.48\textwidth}
    \centering
    \includegraphics[width=\textwidth{}]{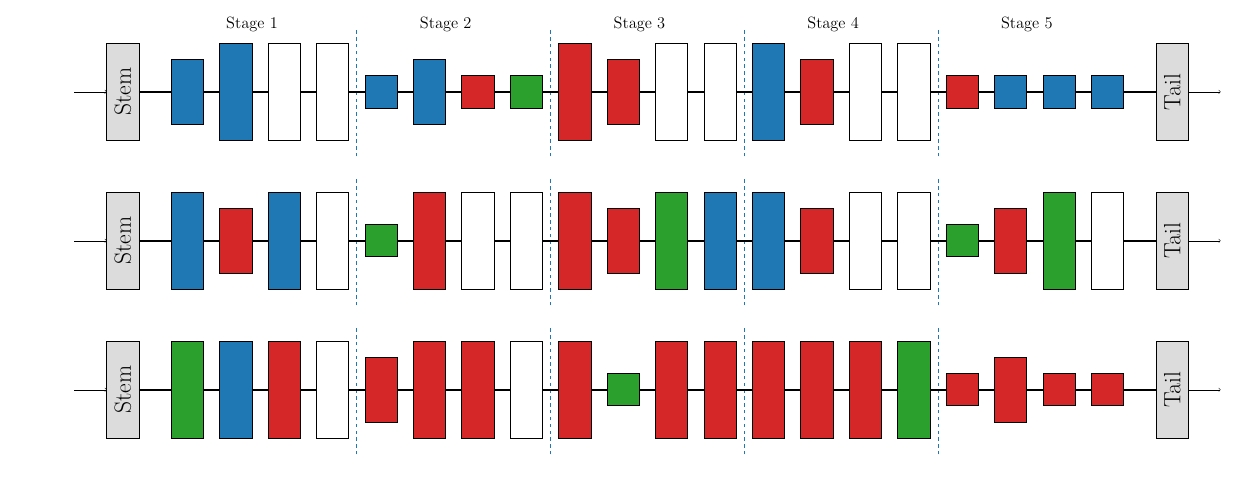}
    \caption{Oxford Flowers-102}
    \end{subfigure}
    \caption{The architectures of \ourmodel{}s referred to in Fig.~1 (main paper) and Fig.~6 (main paper). The \emph{stem} layers in all architectures are the same and not searched. All architectures consist of five blocks, denoted with dashed lines. The first layer in blocks 1, 2, 3, and 5 use stride 2. We use color to denote kernel size and height to denote expansion ratio (legends). For each dataset, architectures are arranged in ascending \#MAdds order from top to bottom, i.e. architectures on the top rows have smaller MAdds than those on the bottom rows.
    \label{fig:archs}}
\end{figure}

\section{Evolved Architectures\label{sec:arch}}
In this section, we visualize the obtained architectures in Fig.~\ref{fig:archs}. All architectures are found by simultaneously maximizing predictive performance and minimizing MAdds. We observe that different datasets require different architectures for an efficient trade-off between MAdds and performance. Finding such architectures is only possible by directly searching on the target dataset, which is the case in \ourmethod{}.

\end{document}